\documentclass[10pt]{article}

\usepackage{amsfonts,amsmath,amsthm}
\usepackage{algorithm,algorithmic}
\usepackage{bm,color}
\usepackage[dvipdfmx]{graphicx} 
\usepackage[margin=1in]{geometry}

\usepackage{mcr}
\usepackage{pkg}
\usepackage{thmE}
\usepackage{algE}

\title{Quick sensitivity analysis for incremental data modification and \\
its application to leave-one-out CV in linear classification problems}

\date{\today}
\author{
Shota Okumura\\
Nagoya Institute of Technology \\
Nagoya, Japan \\
\texttt{okumura.mllab.nit@gmail.com} \\
\and 
Yoshiki Suzuki \\
Nagoya Institute of Technology \\
Nagoya, Japan \\
\texttt{suzuki.mllab.nit@gmail.com} \\
\and 
Ichiro Takeuchi\thanks{Corresponding author} \\
Nagoya Institute of Technology \\
Nagoya, Japan \\
\texttt{takeuchi.ichiro@nitech.ac.jp} \\
}

\begin{document}

\maketitle

\clearpage
\begin{abstract}
We introduce 
a novel sensitivity analysis framework for large scale classification problems 
that can be used when a small number of instances are incrementally added or removed. 
For quickly updating the classifier in such a situation, 
incremental learning algorithms 
have been intensively studied in the literature.
Although they are much more efficient than 
solving the optimization problem from scratch, 
their computational complexity yet 
depends on the entire training set size.
It means that, 
if the original training set is large, 
completely solving an incremental learning problem might be 
still rather expensive.
To circumvent this computational issue, 
we propose a novel framework 
that allows us to make an inference
about the updated classifier
without actually re-optimizing it.
Specifically,
the proposed framework can quickly provide 
a lower and an upper bounds of a quantity 
on the unknown updated classifier. 
The main advantage of the proposed framework 
is that the computational cost of computing these bounds
depends only on the number of updated instances. 
This property is quite advantageous 
in a typical sensitivity analysis task 
where 
only a small number of instances are updated.
In this paper
we demonstrate that
the proposed framework is applicable to 
various practical sensitivity analysis tasks, 
and the bounds
provided by the framework 
are often sufficiently tight for making desired inferences.

\vspace{.1in}
Incremental Learning,
Sensitivity Analysis,
Classification
Support Vector Machines, 
Logistic Regression,
Leave-one-out Cross-validation

\vspace{.1in}
\end{abstract}

\clearpage

\section{Introduction}
 \label{sec:intro}
 Given a large number of training instances, 
 the initial training cost of a classifier 
 such as logistic regression (LR) or support vector machine (SVM)
 would be quite expensive. 
 In principle, 
 there is no simple way around this initial training cost 
 except when suboptimal approximate classifiers 
 (e.g., trained by using random small sub-samples) 
 are acceptable.
 Unfortunately, 
 such an initial training cost 
 is not the only thing
 we must care about in practice. 
 In many practical data engineering tasks, 
 the training set 
 with which the initial classifier was trained 
 might be slightly modified.
 In such a case, 
 it is important to check 
 the \emph{sensitivity}
 of the classifier,
 i.e.,
 how the results would change 
 when the classifier is updated 
 with the slightly modified training set.

Machine learning algorithms
particularly designed
for updating a classifier
when a small number of instances are incrementally added or removed 
are called
\emph{incremental learning} 
\cite{Cauwenberghs2001}. 
For example,
when 
a single instance 
is 
added 
or
removed,
the solution of a linear predictor
can be efficiently computed
(see, e.g., \cite{Hastie2001}).
Incremental learning algorithms 
for SVMs 
and other related learning frameworks 
have been intensively studied 
in the literature
\cite{Cauwenberghs2001,Fine2001,Shilton2005,Laskov2006,Karasuyama2009,Liang2009}.
Even for problems 
whose explicit incremental learning algorithm does not exist, 
\emph{warm start}
approach,
where
the original optimal solution is used as an initial starting point for the updating optimization problem, 
is usually very helpful for reducing incremental learning costs
\cite{Decoste00,Tsai2014}.

However,
the computational cost of incremental learning is still very expensive
if the original training set is large. 
Except for special cases\footnote{
For example, 
in incremental learning of SVM, 
adding
or
removing
an instance
whose margin is greater than one
can be done without any cost
because
such a modification does not change the solution.},
any incremental learning algorithms
must
go through the entire training data matrix at least once,
meaning that 
the complexities
depend on the entire training set size.
When only a small number of instances are modified, 
spending a great amount of computational cost for re-optimizing the classifier
does not seem to be a well worthy effort 
because 
inference results on the updated classifier 
would not be so different
from the original ones.
Furthermore, 
in practical applications, 
it might be computationally intractable
to completely update the classifier 
every time there is a tiny modification of the training set. 
In such a situation, 
it would be nice
if we could quickly check the sensitivity of the classifier 
without actually updating it.
Unless the sensitivity is unacceptably large, 
we might want to use the original classifier
as it is. 

Our key observation here is that 
the goal of sensitivity analysis 
is 
not to update the classifier itself,
but to know
how much
the results of our interest 
would change
when the classifier is updated with the slightly modified training set.
Suppose,
for example,
that we have a test instance. 
Then,
we would be interested in 
whether 
there is a chance that
the class label 
of the test instance 
could be changed
by a minor data modification
or not.
In order to answer such a question, 
we propose a novel approach that can quickly compute 
the sensitivity
of a quantity 
depending on the unknown updated classifier 
without actually re-optimizing it.

In this paper
we study 
a class of regularized linear binary classification problems
with convex loss.
We propose a novel framework 
for this class of problems 
that can efficiently compute 
a lower and an upper bounds
of a general linear score of the updated classifier.
Specifically, 
denoting
the coefficient vector of the updated linear classifier 
as
$\bm \beta^*_{\text{new}}$, 
our framework allows us to obtain
a lower and an upper bounds of
a general linear score
in the form of 
$\bm \eta^\top \bm \beta^*_{\text{new}}$,
where
$\bm \eta$
is an arbitrary vector of the appropriate dimension. 
An advantage of our framework is that
the complexity of computing the bounds
depends only on the number of updated instances,
and does not depend on the size of the entire training set.
This property is quite advantageous
in a typical sensitivity analysis
where 
only a small number of instances are updated.

Bounding a linear score
in the form of
$\bm \eta^\top \bm \beta^*_{\rm new}$
is useful
in a wide range of sensitivity analysis tasks.
First, 
by setting
$\bm \eta = \bm e_j$,
where
$\bm e_j$
is a vector with all 0 except 1 in the $j^{\text{th}}$ position, 
we can obtain 
a lower and an upper bounds
of each coefficient 
$\beta^*_{\text{new}, j}$,
$j = 1, \ldots, d$, 
where
$d$ is the input dimension. 
Another interesting example 
is the case where 
$\bm \eta = \bm x$,
where
$\bm x$
is a test instance of our interest. 
Note that,
if the lower/upper bound of 
$\bm x^\top \bm \beta^*_{\text{new}}$
is positive/negative, 
then we can make sure that the test instance is classified as positive/negative,
respectively.
It means that 
the class label of a test instance might be available 
even if we do not know the exact value of 
$\bm x^\top \bm \beta^*_{\text{new}}$.

To the best of our knowledge,
there are no other existing studies on sensitivity analysis 
that can be used 
as generally as our framework.
However,
there are some closely-related methods 
designed for particular tasks. 
One such example that has been intensively studied in the literature
is
\emph{leave-one-out cross-validation (LOOCV)}.
In each step of an LOOCV, 
a single instance is taken out from the original training set,
and we check 
whether the left-out instance is correctly classified or not
by using the updated classifier. 
This task exactly fits into our framework
because
we are only interested in the class label of the left-out instance,
and the optimal updated classifier itself is not actually required.
Efficient LOOCV methods have been studied 
for SVMs
and other related learning methods
\cite{Jaakkola1999,Joachims2000,Vapnik2000,Zhang2003}\footnote{
In these works,
the main focus is not on computing LOOCV error itself,
but on deriving a lower bound of LOOCV error.}.
Some of these existing methods are built on a similar idea as ours
in the sense that 
the class label 
of a left-out test instance
is efficiently determined
by computing bounds of the linear score
$\bm x^\top \bm \beta^*_{\text{new}}$. 
The bounds obtained by our proposed framework 
are different from 
the bounds used in these existing LOOCV methods.
We empirically show that 
LOOCV computation algorithm using our framework 
is much faster
than existing methods.

The bound computation technique 
we use here is inspired from 
recent studies
on
safe feature screening,
which was introduced in the context of $L_1$ sparse feature modeling
\cite{ElGhaoui2012}.
It allows us
to identify
sparse features
whose coefficients turn out to be zero
at the optimal solution. 
The key idea used there 
is to bound the Lagrange multipliers 
before actually solving the optimization problem for model fitting\footnote{
Lagrange multiplier values at the optimal solution tell us which features are active or non-active.
}.
The idea of bounding the optimal solution
without actually solving the optimization problem 
has been recently extended to various directions 
\cite{ElGhaoui2012,Xiang2011,Ogawa2013,Liu2014,Wang2014,Shibagaki2015}. 
Our main technical contribution
in this paper 
is to
bring this idea to sensitivity analysis problems 
and develop a novel framework 
for efficiently bounding general linear scores 
with the cost depending only on the number of updated instances.

The rest of the paper is organized as follows. 
In \S\ref{sec:setup}, 
we describe the problem setup
and
present three sensitivity analysis tasks
that our framework can be applied to. 
In \S\ref{sec:method} 
we present our main result 
which enables us to compute a lower and an upper bounds of a general linear score 
$\bm \eta^\top \bm \beta^*_{\rm new}$
with the computational cost
depending only on the number of updated instances. 
In addition,
we apply the framework to the three tasks
described in \S\ref{sec:setup}.
In \S\ref{sec:tightening},
we discuss how to tighten the bounds
when the bounds provided by the framework are not sufficiently tight
for making a desired inference.
\S\ref{sec:exp}
is devoted for numerical experiments.
\S~\ref{sec:conc}
concludes the paper and discuss a few future directions of this work. 
All the proofs are presented in Appendix~\ref{app:proof}.

\section{Preliminaries and basic idea}
\label{sec:setup}
In this section 
we first formulate the problem setup
and
clarify the difference 
between
the proposed framework
and 
conventional incremental learning approaches. 
Then,
we discuss three sensitivity analysis tasks
in which
the proposed framework is useful. 

\subsection{Problem setup}
In this paper
we study binary classification problems.
We consider an incremental learning setup, 
where 
we have already trained a classifier 
by using a training set, 
and then 
a small number of instances are
added to
and/or
removed from
the original training set.
The goal of conventional incremental learning problems is to 
update the classifier
by re-training it
with the updated training set.
Hereafter, 
we denote the original and the updated training sets
as
$\{(\bm x_i, y_i)\}_{i \in \cD_{\rm old}}$
and 
$\{(\bm x_i, y_i)\}_{i \in \cD_{\rm new}}$,
respectively, 
where
$\cD_{\rm old}$
and 
$\cD_{\rm new}$
are the set of indices of the instances in old and new training sets
with the sizes 
$n_{\rm old} := |\cD_{\rm old}|$
and 
$n_{\rm new} := |\cD_{\rm new}|$,
respectively.
The input
$\bm x_i$
is assumed to be
$d$-dimensional vector 
and
the class label
$y_i$ 
takes either
$-1$ or $+1$. 
We denote the set of added and removed instances as 
$\{(\bm x_i, y_i)\}_{i \in \cA}$
and
$\{(\bm x_i, y_i)\}_{i \in \cR}$,
where
$\cA \subset \cD_{\rm new}$
and 
$\cR \subset \cD_{\rm old}$
are the set of indices of the added and removed instances
with the sizes 
$n_A := |\cA|$
and 
$n_R := |\cR|$,
respectively.
Note that, 
if one wants to modify an instance in the training set,
one can first remove it 
and
then add the modified one. 

We consider a linear classifier in the form of
\begin{align*}
 \hat{y} = \mycase{
 +1 & \text{if } f(\bm x;\bm \beta) > 0, \\
 -1 & \text{if } f(\bm x;\bm \beta) < 0,
 }
 \text{ with }
 f(\bm x;\bm \beta) = \bm x^\top \bm \beta,
\end{align*}
where the classifier predicts the class label
$\hat{y} \in \{-1, +1\}$
for the given input
$\bm x \in \RR^d$,
while
$\bm \beta \in \RR^d$
is a vector of classifier's coefficients.
In this paper
we consider a class of problems
represented as a minimization of an $L_2$ regularized convex loss. 
Specifically, 
the old and the new classifiers are defined as 
\begin{align}
 \label{eq:prob1}
 \bm \beta^*_{\rm old}
 :=
 \arg \min_{\bm \beta \in \RR^d}
 \frac{1}{n_{\rm old}}
 \sum_{i \in \cD_{\rm old}}
 \ell(y_i, f(\bm x_i; \bm \beta))
 +
 \frac{\lambda}{2}
 \|\bm \beta \|^2,
\end{align}
and
\begin{align}
 \label{eq:prob2}
 \bm \beta^*_{\rm new}
 :=
 \arg \min_{\bm \beta \in \RR^d}
 \frac{1}{n_{\rm new}}
 \sum_{i \in \cD_{\rm new}}
 \ell(y_i, f(\bm x_i; \bm \beta))
 +
 \frac{\lambda}{2}
 \|\bm \beta \|^2,
\end{align}
where the first and the second terms of the objective function
represent
an empirical loss term
and 
an $L_2$-regularization term, 
respectively,
and
$\lambda > 0$
is a regularization parameter
that controls the balance between these two terms.
We assume that
$\ell(\cdot, \cdot)$
is differentiable and convex
with respect to the second argument.
Examples of such a loss function includes 
logistic regression loss
\begin{align}
 \label{eq:logistic-loss}
 \ell(y_i, f(\bm x_i;\bm \beta)) := \log(1 + \exp(- y_i f(\bm x_i;\bm \beta))),
\end{align}
and $L_2$-hinge loss 
\begin{align}
 \label{eq:l2hinge-loss}
 \ell(y_i, f(\bm x_i;\bm \beta)) := \max\{0, 1 - y_i f(\bm x_i;\bm \beta)\}^2.
\end{align}
%
%
%
%
For any $i \in \cD_{\rm old} \cup \cD_{\rm new}$
and 
any  $\bm \beta_0 \in \RR^d$, 
we denote the gradient of the individual loss as 
\begin{align*}
 \bm \nabla \ell_i(\bm \beta_0)
 :=
 \pd{}{\bm \beta} \ell(y_i, f(\bm x_i;\bm \beta))
 \Big|_{\bm \beta = \bm \beta_0}.
\end{align*}

Our main interest is in the cases where 
the number of added instances
$n_A$
and
removed instances
$n_R$
are both much smaller
than the entire training set size 
$n_{\rm old}$
or 
$n_{\rm new}$. 
In such a case,
we expect that
the difference between 
$\bm \beta^*_{\rm old}$
and 
$\bm \beta^*_{\rm new}$ 
is small.
However,
if the training set size
$n_{\rm new}$
is large, 
solving the optimization problem
\eq{eq:prob2}
by using an incremental learning algorithm
is still very expensive 
because any incremental learning algorithms require 
working through the entire training data matrix at least once,
meaning that the complexity of such an incremental learning is at least 
$\cO(n_{\rm new} d)$.

Our approach is different from conventional incremental learning.
In this paper
we propose a novel framework
that enables us to make inferences about the new solution
$\bm \beta^*_{\rm new}$
without actually 
solving the optimization problem
\eq{eq:prob2}.
The proposed framework can efficiently compute 
a lower and an upper bounds of, 
what we call, 
a \emph{linear score}
\begin{align}
 \label{eq:linear-score}
 \bm \eta^\top \bm \beta^*_{\rm new},
\end{align}
where 
$\bm \eta \in \RR^d$
is an arbitrary vector of dimension $d$.
An advantage
of this framework is that
the computational cost of computing these bounds depends
only on the number of updated instances
$n_A + n_R$
and does not depend on
the entire training set size
$n_{\rm old}$
or 
$n_{\rm new}$,
i.e.,
the complexity is 
$\cO((n_A + n_R) d)$.
This property is quite advantageous
in a typical sensitivity analysis 
where 
$n_{\rm new}$
is much larger than 
$n_A + n_R$. 
These bounds are computed
based on the old optimal solution 
$\bm \beta^*_{\rm old}$.
We denote the lower and the upper bounds as
$L(\bm \eta^\top \bm \beta^*_{\rm new})$
and 
$U(\bm \eta^\top \bm \beta^*_{\rm new})$,
respectively,
i.e.,
they satisfy 
\begin{align*}
 L(\bm \eta^\top \bm \beta^*_{\rm new})
 \le
 \bm \eta^\top \bm \beta^*_{\rm new} 
 \le
 U(\bm \eta^\top \bm \beta^*_{\rm new}).
\end{align*}

The proposed framework can be \emph{kernelized} for nonlinear classification problems 
if the inner products 
$\bm \eta^\top \bm \beta^*_{\rm new}$ 
and
$\bm \eta^\top \bm \nabla \ell_i(\bm \beta^*_{\rm old})$ 
for any $i \in \cD_{\rm old} \cup \cD_{\rm new}$
can be represented by the kernel function. 

In the following three subsections,
we discuss three sensitivity analysis tasks
in which
the above proposed framework might be useful. 

\subsection{Sensitivity of coefficients}
\label{subsec:setup-param}
\begin{figure}[t]
 \begin{center}
  \includegraphics[width=0.5\textwidth]{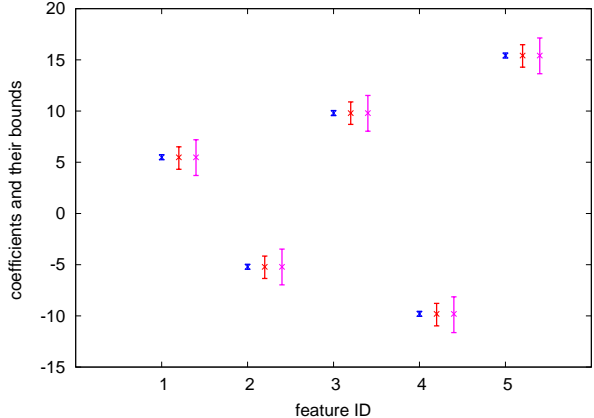} 
  \caption{
  Examples 
  of coefficient bounds 
  $L(\beta^*_{{\rm new},  j})$
  and 
  $U(\beta^*_{{\rm new},  j})$,
  $j \in [5]$,
  for an artificial toy dataset with 
  $n_{\rm old} = 1000$ and $d = 5$.
  The blue, red and pink error bars indicate the bounds 
  when
  $n_A + n_R = 1$ (0.1\%),
  $5$ (0.5\%),
  and
  $10$ (1\%),
  respectively.
  The unknown true coefficients
  $\beta^*_{{\rm new}, j}$,
  $j \in [5]$, 
  are indicated by
  $\bm \times$.
  }
  \label{fig:coef-bound-illustration}
 \end{center}
\end{figure}

Let 
$\bm e_j \in \RR^d$,
$j \in [d]$,
be a vector of all 0 except 1 in the $j^{\rm th}$ element.
Then,
by setting
$\bm \eta := \bm e_j$
in \eq{eq:linear-score},
we can compute
a lower and an upper bounds of the new classifier's coefficient
$\beta^*_{{\rm new}, j} = \bm e_j^\top \bm \beta^*_{\rm new}, j \in [d]$ 
such that
\begin{align*}
 L(\beta^*_{{\rm new}, j}) 
 \le
 \beta^*_{{\rm new}, j} 
 \le
 U(\beta^*_{{\rm new}, j}), 
 ~
 j \in [d].
\end{align*}
Figure~\ref{fig:coef-bound-illustration}
illustrates such coefficient's bounds 
for a simple toy dataset.
Given a lower and an upper bounds of the coefficients,
we can,
in principle, 
obtain the bounds of any quantities 
depending on 
$\bm \beta^*_{\rm new}$. 
Bounding the largest possible change of the new classifier's coefficients 
or a quantity depending on it 
would be beneficial for making decisions in practical tasks.

\subsection{Sensitivity of class labels}
\label{subsec:setup-classification}
\begin{figure}[t]
 \begin{center}
  \includegraphics[width=0.5\textwidth]{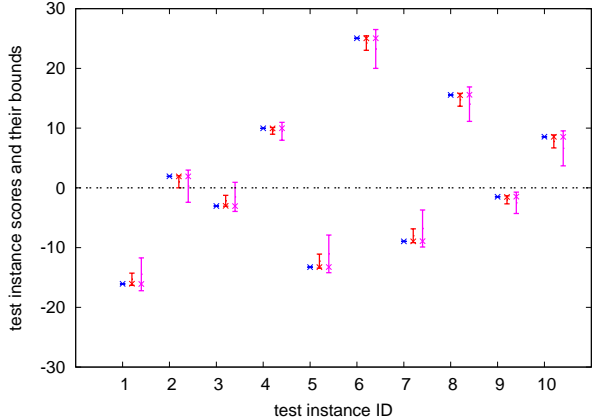} 
  \caption{
  Examples 
  of test instance score bounds 
  $L(\bm x^\top \bm \beta^*_{{\rm new}})$
  and 
  $U(\bm x^\top \bm \beta^*_{{\rm new}})$
  for 10 test instances 
  in the same dataset as in Figure~\ref{fig:coef-bound-illustration}.
  %
  %
  The blue, red and pink error bars indicate the bounds
  when
  $n_A + n_R = 1$ (0.1\%),
  $5$ (0.5\%),
  and
  $10$ (1\%),
  respectively, 
  and 
  the unknown true scores 
  $\bm x^\top \bm \beta^*_{\rm new}$ 
  are indicated by
  $\bm \times$.
  Note that,
  except for the 2nd and the 3rd test instances with $n_A + n_R = 10$ (pink),
  the signs of the lower and the upper bounds are same,
  meaning that the class labels of these test instances are immediately available
  without actually updating the classifier. 
  }
  \label{fig:classification-bound-illustration}
 \end{center}
\end{figure}

Next,
let us consider
sensitivity analysis of the new class label for a test instance 
$\bm x \in \RR^d$,
i.e.,
we would like to know 
\begin{align*}
 \hat{y}
 :=
 {\text{sgn}}(f(\bm x; \bm \beta^*_{\rm new}))
 =
 {\text{sgn}}(\bm x^\top \bm \beta^{*}_{\rm new}).
\end{align*}
By setting
$\bm \eta := \bm x$
in \eq{eq:linear-score},
we can compute a lower and an upper bounds 
such that 
\begin{align}
 \label{eq:classification-bounds}
 L(\bm x^\top \bm \beta^*_{\text{new}})
 \le 
 \bm x^\top \bm \beta^*_{\text{new}}
 \le 
 U(\bm x^\top \bm \beta^*_{\text{new}}). 
\end{align}
Here,
it is interesting to note that,
using the following simple facts: 
\begin{subequations}
 \label{eq:classification-facts}
\begin{align}
 &
 L(\bm x^\top \bm \beta^*_{\text{new}}) \ge 0
 ~\Rightarrow~
 \hat{y} = +1,
 \\
 &
 U(\bm x^\top \bm \beta^*_{\text{new}}) < 0
 ~\Rightarrow~
 \hat{y} = -1, 
\end{align}
\end{subequations}
the class label 
$\hat{y}$
can be available
without actually obtaining
$\bm \beta^*_{\text{new}}$ 
if 
the bounds are sufficiently tight 
such that the signs of the lower and the upper bounds are same.
If the number of updated instances
$n_A + n_R$
is relatively smaller than
the entire training set size
$n_{\rm old}$
or
$n_{\rm new}$, 
we expect that the two solutions
$\bm \beta^*_{\rm old}$ 
and 
$\bm \beta^*_{\rm new}$
would not be so different. 
In such cases, 
as we demonstrate empirically in \S\ref{sec:exp}, 
the bounds
in \eq{eq:classification-bounds}
are sufficiently tight 
in many cases. 
Figure~\ref{fig:classification-bound-illustration}
illustrates the tightness of the bounds 
in a toy dataset. 

\subsection{Leave-one-out cross-validation (LOOCV)}
\label{subsec:setup-loocv}
One of the traditional problem setups to which our proposed framework
can be naturally applied is leave-one-out cross-validation (LOOCV).
The LOOCV error is defined as
\begin{align*}
 \text{LOOCV error}
 :=
 \frac{1}{n} 
 \sum_{h \in [n]}
 I(
 y_h \neq \text{sgn}(\bm x^\top_i \bm \beta^*_{(-h)})
 ),
\end{align*}
where
${\rm sgn}(\cdot)$
is the sign,
and 
$\bm \beta^*_{(-h)}$
is the optimal solution after leaving out the $h^{\rm th}$ instance,
which is defined as 
\begin{align*}
 \bm \beta^*_{(-h)}
 :=
 \arg \min_{\bm \beta \in \RR^d}
 ~
 \frac{1}{n_{\rm old} - 1}
 \sum_{i \in \cD_{\rm old} \setminus \{h\}}
 \ell(y_i, f(\bm x_i, \bm \beta))
 +
 \frac{\lambda}{2}
 \|\bm \beta\|^2. 
\end{align*}
Here,
our idea is to regard the solution
obtained by the whole training set as 
$\bm \beta^*_{\text{old}}$,
and 
$\bm \beta^*_{(-h)}$
as
$\bm \beta^*_{\text{new}}$.
By setting
$\bm \eta := y_h \bm x_h$
in
\eq{eq:linear-score},
we can compute the bounds 
such that 
 \begin{align}
  \label{eq:loocv-bounds}
  L(y_h \bm x^\top_h \bm \beta^*_{(-h)})
  \le
  y_h \bm x^\top_h \bm \beta^*_{(-h)}
  \le
  U(y_h \bm x^\top_h \bm \beta^*_{(-h)}).
 \end{align}
 These bounds
 in 
 \eq{eq:loocv-bounds} 
 can be used to know
 whether the left out instance is correctly classified or not.
 If the lower bound is positive,
 the left-out instance will be correctly classified,
 while
 it will be mis-classified 
 if the upper bound is negative.

 Using
 \eq{eq:loocv-bounds},
 we can also obtain the bounds on the LOOCV error itself: 
 \begin{align*}
  \text{LOOCV error}
  &
  \ge 
  \frac{1}{n}
  \sum_{h \in [n]}
  I\left(
  U(y_h \bm x^\top_h \bm \beta^*_{(-h)}) < 0
  \right),
  \\
  \text{LOOCV error}
  &
  \le 
  1 - 
  \frac{1}{n}
  \sum_{h \in [n]}
  I\left(
  L(y_h \bm x^\top_h \bm \beta^*_{(-h)}) > 0
  \right),
 \end{align*}
 where
 $I(\cdot)$
 is the indicator function.
In numerical experiments,
we illustrate that
this approach works quite well.
 \section{Quick sensitivity analysis}
\label{sec:method}
In this section
we present our main results
on our quick sensitivity analysis framework.
The following theorem tells that
we can compute 
a lower and an upper bounds of
a general linear score
$\bm \eta^\top \bm \beta^*_{\text{new}}$
by using the original solution 
$\bm \beta^*_{\text{old}}$. 

\begin{theo}
 \label{theo:main}
 Let
 \begin{align}
  \label{eq:delta-s}
  \bm \Delta s
  :=
  \frac{1}{n_A + n_R}
  \left(
  \sum_{i \in \cA}
  \bm \nabla \ell_i(\bm \beta^*_{\text{old}})
  - 
  \sum_{i \in \cR}
  \bm \nabla \ell_i(\bm \beta^*_{\text{old}})
  \right).
 \end{align}
 Then,
 for an arbitrary vector
 $\bm \eta \in \RR^d$,
 the linear score
 $\bm \eta^\top \bm \beta^*_{\text{new}}$
 satisfies 
 \begin{subequations}
  \label{eq:main-bounds}
  \begin{align}
   \nonumber
   &
   \bm \eta^\top \bm \beta^*_{\text{new}}
   \ge
   L(\bm \eta^\top \bm \beta^*_{\text{new}})
   \\
   &
   ~~~~~
   :=
   \frac{n_{\text{new}} + n_{\text{old}}}{2n_{\text{new}}}
   \bm \eta^\top \bm \beta^*_{\text{old}}
   -
   \lambda^{-1}
   \frac{n_A + n_R}{2n_{\text{new}}} 
   \bm \eta^\top \bm \Delta s
   \label{eq:main-LB}
   -
   \frac{1}{2}
   \|\bm \eta\|
   \left\|
   \frac{n_A - n_R}{n_{\text{new}}} \bm \beta^*_{\text{old}}
   +
   \lambda^{-1} \frac{n_A + n_R}{n_{\text{new}}} \bm \Delta s
   \right\|,
  \end{align}
  \begin{align}
   \nonumber
   &
   \bm \eta^\top \bm \beta^*_{\text{new}}
   \le
   U(\bm \eta^\top \bm \beta^*_{\text{new}})
   \\
   &
   ~~~~~
   :=
   \frac{n_{\text{new}} + n_{\text{old}}}{2n_{\text{new}}}
   \bm \eta^\top \bm \beta^*_{\text{old}}
   -
   \lambda^{-1}
   \frac{n_A + n_R}{2n_{\text{new}}} 
   \bm \eta^\top \bm \Delta s
   \label{eq:main-UB}
   +
   \frac{1}{2}
   \|\bm \eta\|
   \left\|
   \frac{n_A - n_R}{n_{\text{new}}} \bm \beta^*_{\text{old}}
   +
   \lambda^{-1} \frac{n_A + n_R}{n_{\text{new}}} \bm \Delta s
   \right\|.
  \end{align}
 \end{subequations}
\end{theo}
The proof is presented in Appendix~\ref{app:proof}.
%

An advantage of the bounds in 
\eq{eq:main-bounds}
is that the computational complexity 
does not depend on
the total number of instances,
but only on the number of modified instances.
%
It is easy to confirm that
the main computational cost of these bounds 
is
in the computation of 
$\bm \Delta s$
in \eq{eq:delta-s},
and 
its complexity is 
$\cO((n_A + n_R)d)$.
The tightness of the bounds,
i.e.,
the difference between the upper and the lower bounds is written as
\begin{align}
 U(\bm \eta^\top \bm \beta^*_{\text{new}})
 -
 L(\bm \eta^\top \bm \beta^*_{\text{new}})
 \label{eq:tightness}
 =
 \|\bm \eta\|
 \left\|
 \frac{n_A - n_R}{n_{\text{new}}} \bm \beta^*_{\text{old}} + \lambda^{-1} \frac{n_A + n_R}{n_{\text{new}}} \bm \Delta s
 \right\|.
\end{align}
In a typical sensitivity analysis
where
$n_A$
and 
$n_R$
are much smaller than
$n_{\text{new}}$, 
the tightness in
\eq{eq:tightness}
would be small. 
Note also that the tightness depends inversely on the regularization parameter
$\lambda$.
If
$\lambda$
is very small and close to zero,
the bounds become very loose.
%


\subsection{Sensitivity analysis of coefficients}
\label{subsec:main-param-sensitivity}
As discussed in
\S\ref{subsec:setup-param},
by substituting 
$\bm \eta := \bm e_j$,
$j \in [d]$,
into 
\eq{eq:main-bounds},
we obtain a lower and an upper bounds of 
the
$j^{\rm th}$
coefficient of the new classifier.

\begin{coro}
\label{coro:param-bound}
 For
 $j \in [d]$, 
 the
 $j^{\rm th}$
 coefficient
 of the new classifier 
 satisfies
 \begin{subequations}
  \label{eq:param-bound}
  \begin{align}
   \nonumber
   &
   \beta^*_{{\rm new}, j}
   \ge
   L(\beta^*_{{\rm new}, j})
   \\
   &
   ~~~~~
   :=
   \frac{
   n_{\rm new} + n_{\rm old}
   }{
   2 n_{\rm new}
   }
   \beta^*_{{\rm old}, j}
   -
   \lambda^{-1}
   \frac{
   n_A + n_R
   }{
   2 n_{\rm new}
   }
   \Delta s_j
   \label{eq:param-LB}
   -
   \frac{1}{2}
   \Big\|
   \frac{
   n_A - n_R
   }{
   n_{\rm new}
   }
   \bm \beta^*_{{\rm old}}
   +
   \lambda^{-1}
   \frac{
   n_A + n_R
   }{
   n_{\rm new}
   }
   \bm \Delta s
   \Big\|,
  \end{align}
  \begin{align}
   \nonumber
   &
   \beta^*_{{\rm new}, j}
   \le
   U(\beta^*_{{\rm new}, j})
   \\
   &
   ~~~~~
   :=
   \frac{
   n_{\rm new} + n_{\rm old}
   }{
   2 n_{\rm new}
   }
   \beta^*_{{\rm old}, j}
   +
   \lambda^{-1}
   \frac{
   n_A + n_R
   }{
   2 n_{\rm new}
   }
   \Delta s_j
   \label{eq:param-LB}
   +
   \frac{1}{2}
   \Big\|
   \frac{
   n_A - n_R
   }{
   n_{\rm new}
   }
   \bm \beta^*_{{\rm old}}
   +
   \lambda^{-1}
   \frac{
   n_A + n_R
   }{
   n_{\rm new}
   }
   \bm \Delta s
   \Big\|.
  \end{align}
 \end{subequations}
\end{coro}

Note that the third term does not depend on $j \in [d]$, i.e., 
the tightness of the bounds in 
\eq{eq:param-bound}
is common for all the coefficients 
$\beta^*_{{\rm new}, j}$,
$j \in [d]$.

Given a lower and an upper bounds of the coefficients 
$\beta^*_{{\rm new}, j}$,
$j \in [d]$, 
we can obtain the bounds of any quantities 
depending on 
$\bm \beta^*_{\rm new}$. 
For example,
it is straightforward to know 
how much the classifier's coefficients can change by the incremental operation
when the amount of the change is measured in terms of some norm of
$\bm \beta^*_{\rm new} - \bm \beta^*_{\rm old}$. 
%
\begin{coro}
 For any
 $q > 0$,
 let
 $\|\bm z\|_q$
 be the
 $L_q$ 
 norm of a vector
 $\bm z$.
 Then the difference
 between
 the old and the new classifier's coefficients 
 in $L_q$-norm 
 is bounded from above as 
 \begin{align}
  \label{eq:norm-bound}
  \| \bm \beta^*_{\rm new} - \bm \beta^*_{\rm old}\|_q
  \le
  \Bigl(
  \sum_{j \in [d]}
  \max
  \{
  \beta^*_{{\rm old}, j} - L(\beta^*_{{\rm new}, j}), 
  U(\beta^*_{{\rm new}, j}) - \beta^*_{{\rm old}, j}
  \}^q
  \Bigr)^{\frac{1}{q}}.
 \end{align}
\end{coro}
%


Some readers might note that a lower and an upper bounds of a general linear score 
$\bm \eta^\top \bm \beta^*_{\rm new}$
can be simply
obtained 
by using the bounds of each coefficients
$\beta^*_{{\rm new}, j}$,
$j \in [d]$. 
%
%
Such naive bounds are given as 
\begin{subequations} 
 \label{eq:naive-bound}
 \begin{align}
  &
  \bm \eta^\top \bm \beta^*_{\rm new}
  \ge
  \tilde{L}(\bm \eta^\top \bm \beta^*_{\rm new})
  :=
  \sum_{j | \eta_j < 0}
  \eta_j U(\beta^*_{{\rm new}, j})
  +
  \sum_{j | \eta_j > 0}
  \eta_j L(\beta^*_{{\rm new}, j}),
  \\
  &
  \bm \eta^\top \bm \beta^*_{\rm new}
  \le
  \tilde{U}(\bm \eta^\top \bm \beta^*_{\rm new})
  :=
  \sum_{j | \eta_j < 0}
  \eta_j L(\beta^*_{{\rm new}, j})
  +
  \sum_{j | \eta_j > 0}
  \eta_j U(\beta^*_{{\rm new}, j}).
  \end{align}
\end{subequations}
The tightness of the bounds in
\eq{eq:naive-bound}
is written as
\begin{align*}
 \tilde{U}(\bm \eta^\top \bm \beta^*_{\rm new})
 -
 \tilde{L}(\bm \eta^\top \bm \beta^*_{\rm new})
 :=
 \sum_{j \in [d]}
 |\eta_j|
 (U(\beta^*_{{\rm new}, j}) - L(\beta^*_{{\rm new}, j})),
 \end{align*}
 which is clearly much looser than
 \eq{eq:tightness}.
 Thus,
 if the quantity of the interest is written in a linear score form
 $\bm \eta^\top \bm \beta^*_{\rm new}$,
 we should use the bounds
 in
 \eq{eq:main-bounds}
 rather than
 \eq{eq:naive-bound}.

\subsection{Sensitivity analysis of class labels}
\label{subsec:main-classification-sensitivity}
Next,
we use 
Theorem~\ref{theo:main}
for sensitivity analysis of new class labels.
As discussed in
\S\ref{subsec:setup-classification},
for an input vector
$\bm x \in \RR^d$,
we can obtain a lower and an upper bounds of a linear score 
$\bm x^\top \bm \beta^*_{\rm new}$
by setting 
$\bm \eta = \bm x$.
From 
\eq{eq:classification-facts}, 
we can know the new class label 
if the signs of the lower and the upper bounds are same. 

\begin{coro}
 \label{coro:classification-sensitivity}
 Let
 $\bm x \in \RR^d$
 be an arbitrary $d$-dimensional input vector.
 Then,
 the classification result
 \begin{align*}
  \hat{y}
  :=
  {\rm sgn}(f(\bm x ; \bm \beta^*_{\rm new}))
  =
  {\rm sgn}(\bm x^\top \bm \beta^*_{\rm new})
 \end{align*}
 satisfies 
 \begin{align*}
  \hat{y}
  =
  \mycase{
  +1
  &
  \text{if }
  L(\bm x^\top \bm \beta^*_{\rm new}) > 0,
  \\
  -1
  &
  \text{if }
  U(\bm x^\top \bm \beta^*_{\rm new}) < 0,
  \\
  \text{unknown}
  &
  \text{otherwise},
  }
 \end{align*}
 where
 \begin{align*}
  L(\bm x^\top \bm \beta^*_{\rm new})
  :=
  \frac{n_{\text{new}} + n_{\text{old}}}{2n_{\text{new}}}
  \bm x^\top \bm \beta^*_{\text{old}}
  -
  \lambda^{-1}
  \frac{n_A + n_R}{2n_{\text{new}}} 
  \bm x^\top \bm \Delta s
  -
  \frac{1}{2}
  \|\bm x\|
  \left\|
  \frac{n_A - n_R}{n_{\text{new}}} \bm \beta^*_{\text{old}}
  +
  \lambda^{-1} \frac{n_A + n_R}{n_{\text{new}}} \bm \Delta s
  \right\|,
 \end{align*}
 \begin{align*}
  U(\bm x^\top \bm \beta^*_{\rm new})
  :=
  \frac{n_{\text{new}} + n_{\text{old}}}{2n_{\text{new}}}
  \bm x^\top \bm \beta^*_{\text{old}}
  -
  \lambda^{-1}
  \frac{n_A + n_R}{2n_{\text{new}}} 
  \bm x^\top \bm \Delta s
  +
  \frac{1}{2}
  \|\bm x\|
  \left\|
  \frac{n_A - n_R}{n_{\text{new}}} \bm \beta^*_{\text{old}}
  +
  \lambda^{-1} \frac{n_A + n_R}{n_{\text{new}}} \bm \Delta s
  \right\|.
 \end{align*}
\end{coro}

Corollary~\ref{coro:classification-sensitivity} 
is useful in transductive setups~\cite{Vapnik96}
where
we are only interested in the class labels 
of the prespecified set of test inputs. 
%


\subsection{Quick leave-one-out cross-validation}
\label{subsec:main-LOOCV}
In LOOCV,
we repeat
leaving out a single instance
from the training set,
and check
whether it is correctly classified or not
by the new classifier
which is trained without the left-out instance. 
Thus,
each step of LOOCV computation
can be considered as an incremental operation
with
$n_A = 0$
and
$n_R = 1$.
%
%
Denoting the left-out instance as 
$(\bm x_h, y_h)$,
$h \in [n_{\rm old}]$, 
the task is to inquire whether the left-out instance is correctly classified or not,
which is known by checking the sign of
$y_h f(\bm x_h; \bm \beta^*_{\rm new}) = y_h \bm x_h^\top \bm \beta^*_{\rm new}$. 

\begin{coro}
 Consider a single step of LOOCV computation
 where
 an instance
 $(\bm x_h, y_h), h \in [n_{\rm old}]$,
 is left out.
 Then,
 \begin{align*}
  &
  L(y_h f(\bm x_h; \bm \beta^*_{\rm new})) > 0
  ~\Rightarrow~
  \text{$(\bm x_h, y_h)$ is correctly classified}
  \\
  &
  U(y_h f(\bm x_h; \bm \beta^*_{\rm new})) < 0
  ~\Rightarrow~
  \text{$(\bm x_h, y_h)$ is mis-classified}
 \end{align*}
 where
 \begin{subequations} 
  \label{eq:bounds-LOOCV}
 \begin{align}
 \nonumber  
 &
 L(y_h f(\bm x_h; \bm \beta^*_{\rm new}))
 \\
  &
 :=
  \frac{2 n_{\rm old} - 1}{2n_{\rm old} - 2}
  y_h \bm x_h^\top \bm \beta^*_{\text{old}}
 +
 \frac{\lambda^{-1}}{2n_{\text{old}} - 2}
  y_h \bm x_h^\top \bm \nabla \ell_i(\bm \beta^*_{\rm old})
  \label{eq:L-bounds-LOOCV}
  -
  \frac{1}{2}
  \|\bm x_h\|
  \left\|
  \frac{-1}{n_{\rm old} - 1} \bm \beta^*_{\text{old}}
  +
 \frac{\lambda^{-1}}{n_{\rm old} - 1} \bm \nabla \ell_i(\bm \beta^*_{\rm old})
  \right\|,
 \end{align}
\begin{align}
 \nonumber
 &
 U(y_h f(\bm x_h; \bm \beta^*_{\rm new}))
 \\
 &
 :=
  \frac{2 n_{\rm old} - 1}{2n_{\rm old} - 2}
  y_h \bm x_h^\top \bm \beta^*_{\text{old}}
 +
 \frac{\lambda^{-1}}{2n_{\text{old}} - 2}
  y_h \bm x_h^\top \bm \nabla \ell_i(\bm \beta^*_{\rm old})
  \label{eq:U-bounds-LOOCV}
 +
  \frac{1}{2}
  \|\bm x_h\|
  \left\|
  \frac{-1}{n_{\rm old} - 1} \bm \beta^*_{\text{old}}
  +
 \frac{\lambda^{-1}}{n_{\rm old} - 1} \bm \nabla \ell_i(\bm \beta^*_{\rm old})
  \right\|.
 \end{align}
 \end{subequations}
\end{coro}

\section{Tightening linear score bounds via a suboptimal solution}
\label{sec:tightening}
In the previous section
we introduced a framework
that can quickly compute
a lower and an upper bounds of
a linear score of the new classifier.
Unfortunately,
it is not always the case that
these bounds are sufficiently tight 
for making a desired inference on the new classifier. 
For example, 
if the lower and the upper bounds of
$\bm x^\top \bm \beta_{\rm new}^*$
do not have the same sign
for a test input
$\bm x$,
we cannot tell
which class it would be classified to.
In this section
we discuss how to deal with such a situation.

The simplest way
to handle such a situation 
is just to use
conventional incremental learning algorithms.
If we completely solve the optimization problem
\eq{eq:prob2}
by an incremental learning algorithm,
we can obtain 
$\bm \beta_{\rm new}^*$
itself.
However,
if our goal is only to make a particular inference about the new classifier, 
we do not have to solve the optimization problem
\eq{eq:prob2}
completely until convergence. 
In this section 
we propose a similar framework 
for computing 
a lower and an upper bounds
of a linear score
by using a suboptimal solution
before convergence 
which would be obtained
during the optimization of problem 
\eq{eq:prob2}.

We denote such a suboptimal solution as
$\hat{\bm \beta}_{\rm new}$.
In order to compute the bounds,
we use the gradient information of the problem \eq{eq:prob2},
which we denote
\begin{align}
 \bm g(\hat{\bm \beta}_{\rm new})
 :=
 \frac{1}{n_{\rm new}}
 \sum_{i \in \cD_{\rm new}}
 \bm \nabla \ell_i(\hat{\bm \beta}_{\rm new})
 +
 \lambda
 \hat{\bm \beta}_{\rm new}.
\end{align}
%
The complexity of 
computing the gradient vector 
from scratch 
is 
$\cO(n_{\rm new} d)$.
However, 
if we are using a gradient-based optimization algorithm 
such as
conjugate gradient or quasi-Newton methods, 
we should have already computed the gradient vector
in each iteration of the optimization algorithm.
The following theorem 
provides
a lower and an upper bound
of a linear score
by using the current gradient information.
If we already have computed
$\bm g(\hat{\bm \beta}_{\rm new})$,
these bounds can be obtained very cheaply. 
%
\begin{theo}
 \label{theo:grad} 
 For an arbitrary vector
 $\bm \eta \in \RR^d$,
 the linear score
 $\bm \eta^\top \bm \beta^*_{\text{new}}$
 satisfies 
 \begin{subequations} 
  \label{eq:grad-bound}
  \begin{align}
   \label{eq:grad-LB}
   \bm \eta^\top \bm \beta^*_{\text{new}}
   \ge
   \hat{L}(\bm \eta^\top \bm \beta^*_{\text{new}})
   := \bm \eta^\top \hat{\bm \beta}_{\rm new}
   - \frac{\lambda^{-1}}{2} \bm \eta^\top \bm g(\hat{\bm \beta}_{\rm new})
   - \frac{\lambda^{-1}}{2}
   \|\bm \eta\| \|\bm g(\hat{\bm \beta}_{\rm new})\|,
  \end{align}
  \begin{align}
   \label{eq:grad-UB}
   \bm \eta^\top \bm \beta^*_{\text{new}}
   \le
   \hat{U}(\bm \eta^\top \bm \beta^*_{\text{new}})
   := \bm \eta^\top \hat{\bm \beta}_{\rm new}
   - \frac{\lambda^{-1}}{2} \bm \eta^\top \bm g(\hat{\bm \beta}_{\rm new})
   + \frac{\lambda^{-1}}{2}
   \|\bm \eta\| \|\bm g(\hat{\bm \beta}_{\rm new})\|.
  \end{align}
 \end{subequations}
\end{theo}
The proof is presented in Appendix~\ref{app:proof}.

A nice property of the bounds in
\eq{eq:grad-bound}
is that the tightness 
\begin{align*}
 \hat{U}(\bm \eta^\top \bm \beta^*_{\text{new}})
 -
 \hat{L}(\bm \eta^\top \bm \beta^*_{\text{new}})
 =
 \lambda^{-1} \|\bm \eta\| \|\bm g(\hat{\bm \beta}_{\rm new})\|
\end{align*}
is linear in the norm of the gradient
$\|\bm g(\hat{\bm \beta}_{\rm new})\|$.
It means that,
as the optimization algorithm for
\eq{eq:prob2}
proceeds,
the gap between the lower and the upper bounds
in \eq{eq:grad-bound}
decreases, 
and
it converges to zero 
as the solution converges to the optimal one.
Theorem~\ref{theo:grad} 
can be used 
as a stopping criterion 
for incremental learning optimization problem
\eq{eq:prob2}.
For example,
in a sensitivity analysis of class labels,
one can proceed the optimization process
until
the signs of the lower and the upper bounds
in
\eq{eq:grad-bound}
become same.
 \section{Numerical experiments}
\label{sec:exp}
In this section we describe numerical experiments. 
%
In 
\S\ref{subsec:exp-QSA} 
we illustrate the tightness and the computational efficiency 
of our bounds 
in two sensitivity analysis tasks
described in 
\S\ref{subsec:setup-param}
and 
\S\ref{subsec:setup-classification}.
In 
\S\ref{subsec:exp-LOOCV} 
we apply our framework to LOOCV computation 
as described in 
\S\ref{subsec:setup-loocv} 
and 
compare its performance 
with conventional LOOCV computation methods.

Table~\ref{tab:datasets}
summarizes the datasets
used in the experiments.
They are all taken from
libsvm dataset repository
\cite{Chang2011a}.
For the experiments in \S\ref{subsec:exp-QSA},
we used larger datasets {\tt D5}-{\tt D8}.
For LOOCV experiments in \S\ref{subsec:exp-LOOCV},
we used smaller datasets {\tt D1}-{\tt D4}.
As examples of the loss function
$\ell$,
we used
LR loss
\eq{eq:logistic-loss}
and
SVM loss
\eq{eq:l2hinge-loss}.
In \S\ref{subsec:exp-QSA} 
we only show the results on logistic regression.
%
In \S\ref{subsec:exp-LOOCV}
we compare our results on SVMs with conventional LOOCV methods particularly designed for SVMs.
For logistic regression,
we only compare our framework with
conventional incremental learning algorithm 
because
there is no particular LOOCV computation method for logistic regression.
As an incremental learning algorithm, 
which is used as competitor and as a part of our algorithm for LOOCV computation, 
we used the approach in \cite{Tsai2014}.
All the computations were conducted by using a single core of an HP workstation
Z820 (Xeon(R) CPU E5-2693 (3.50GHz), 64GB MEM).

\begin{table}[t]
 \begin{center}
  \caption{Benchmark datasets used in the experiments.}
  \label{tab:datasets}
  \begin{scriptsize}
\begin{tabular}{|c||c|c|c|c|c|}
\hline
&dataset name & $n_{\rm train}$ & $d$ & $n_{\rm test}$\\
\hline
\hline
D1&sonar&208 &60&not used\\
\hline
D2&splice&1000&60&not used\\
\hline
D3&w5a&9888&300&not used\\
\hline
D4&a7a&16100&123&not used\\
\hline
D5&a9a&32561&123&16281\\
\hline
D6&ijcnn&49990&22&91701\\
\hline 
D7&cod-rna&59535&8&271617\\
\hline
D8&kdd2010&$>$ 8 million &$>$20 million & $>$ 0.5 million\\
 \hline
\end{tabular}
  \end{scriptsize}
 \end{center}
\end{table}

\subsection{Results on two sensitivity analysis tasks}
\label{subsec:exp-QSA}
Here 
we show the results on two sensitivity analysis tasks
described in
\S\ref{subsec:setup-param}
and 
\S\ref{subsec:setup-classification}.
We empirically evaluate 
the tightness of the bounds 
and
the computational costs
for larger datasets
{\tt D5}-{\tt D8}.
First, 
we see how the results change 
as the number of added and/or removed instances changes among 
$n_A + n_R \in \{0.01\%$,
$0.02\%$,
$0.05\%$,
$0.1\%$,
$0.2\%$,
$0.5\%$,
$1\%\}$
of the entire training set size 
$n_{\rm old}$.
Next,
we see the results 
when the number of the entire training set size changes among
$n_{\rm old} \in \{10\%$,
$20\%$,
$\ldots$,
$90\%$,
$99\%\}$
of
$n_{\rm train}$, 
while the number of added and/or removed instances is fixed to 
$n_A + n_R = 0.001n_{\rm train}$.

In the first sensitivity analysis task
about coefficients
(see \S\ref{subsec:setup-param} and \S\ref{subsec:main-param-sensitivity}), 
we simply computed 
the difference between the upper and the lower bounds
$U(\beta_{{\rm new}, j}^*) - L(\beta_{{\rm new}, j}^*)$
for evaluating the tightness of the bounds.
%
For the second sensitivity analysis task
about class labels 
(see \S\ref{subsec:setup-classification} and \S\ref{subsec:main-classification-sensitivity}),
we examined the percentage of the test instances 
for which 
the signs of the lower and the upper bounds are same.
Remember that 
the class label can be immediately available 
when the lower and the upper bounds have same sign.

Tables~\ref{tab:exp-result3} - \ref{tab:exp-result8}
show the results 
for the former task.
(Figure~\ref{fig:exp-result1} depicts the results on {\tt D8} as an example).
%
%
%
%
%
These results indicate that the bounds are fairly tight 
if the 
$n_A + n_R$
is relatively smaller than
$n_{\rm old}$.
The computational costs of our proposed framework 
(blue thick curves) 
are negligible
compared with the costs of actual incremental learning
(red thick curves). 

Tables~\ref{tab:exp-result9} - \ref{tab:exp-result14}
show the results for the latter task
(Figure~\ref{fig:exp-result2} depicts the results on {\tt D8} as an example).
%
%
%
The results here indicate that,
in most cases,
the bounds are sufficiently tight
for making the signs of the lower and the upper bounds same. 
It means that,
in most cases,
the new class labels after incremental operation
are available 
without actually updating the classifier itself.

The results presented here were obtained
with the regularization parameter
$\lambda = 0.01, 0.1, 1$.
We observed that, 
for larger $\lambda$,
the bounds became tighter.
These all experiment results are mean and variance of performing 30 times.

\begin{footnotesize}
 \begin{algorithm}[t]
  \caption{Proposed LOOCV method (op1)}
  \label{alg:loocv}
  \begin{algorithmic}[1]
   \REQUIRE $\{(\bm x_i, y_i)\}_{i \in [n_{\rm old}]}$
   \STATE $\bm \beta_{\rm all}^{*}\leftarrow $solve \eq{eq:prob1}, $h \leftarrow 1$, $err \lA 0$
   \WHILE {$h \leq n_{\rm old}$}
   \IF {$U(y_h f(\bm x_h; \bm \beta^*_{(-h)}) < 0$}
   \STATE $err \lA err + 1$
   \ELSIF {$L(y_h f(\bm x_h; \bm \beta^*_{(-h)}) < 0$}
   \STATE $\bm \beta_{(-h)}^{*}\leftarrow $solve \eq{eq:prob2} by incremental learning algorithm
   \IF {$y_h \bm x_h^\top \bm \beta^*_{(-h)} < 0$}
   \STATE $err \lA err + 1$
   \ENDIF
   \ENDIF
   \STATE $h \lA h + 1$
   \ENDWHILE
   \ENSURE LOOCV error: $err/n_{\rm old}$
  \end{algorithmic}
 \end{algorithm}
\end{footnotesize}

 \subsection{Leave-one-out cross-validation}
\label{subsec:exp-LOOCV}
We applied the proposed framework
to LOOCV task,
and compare its computational efficiency
with existing approaches.
We consider two options.
In the first option ({\bf op1}),
we only used the method described in \S\ref{subsec:main-LOOCV}.
In the second option ({\bf op2}), 
we also used the method described in \S\ref{sec:tightening}.
Algorithm~\ref{alg:loocv}
is the pseudo-code 
for computing LOOCV errors 
by using the proposed framework with {\bf op1}. 
Briefly speaking, 
for each of the left-out instance
$(\bm x_h, y_h)$, 
we first compute the lower and the upper bounds of
$y_h f(\bm x_h; \bm \beta^*_{(-h)})$.
Then,
if the lower/upper bound is greater/smaller than zero,
we confirm that the instance is correctly/incorrectly classified.
When the signs of the two bounds are different,
the class label is unknown.
In such a case, 
we use conventional incremental learning algorithm in \cite{Tsai2014}.
In op1, 
we ran the incremental learning algorithm
until its convergence
and
obtain 
$\bm \beta^*_{(-h)}$
itself.
In op2,
we stopped the optimization process 
at which the signs of the bounds in
\eq{eq:grad-bound}
became same.

For SVMs, 
several LOOCV methods
have been studied
in the literature~\cite{Vapnik96,Joachims2000}. 
For the experiments with SVM loss,
we thus compare our approach with the methods in
\cite{Vapnik96} and \cite{Joachims2000}.
The former approach merely exploits the fact 
that adding and/or removing non-support vectors 
does not change the classifier.
The method called
{$\xi\text{-}\alpha$ estimator}
\cite{Joachims2000}
also provides a lower bound of 
$y_h f(\bm x_h; \bm \beta^*_{\rm new})$ 
without actually obtaining 
$\bm \beta^*_{\rm new}$.
%
%
For the experiments with logistic regression loss, 
we compare our approaches only with incremental learning approach \cite{Tsai2014} 
because 
there are no other competing methods.

We used the above LOOCV computation methods
in 
model selection tasks
for linear and nonlinear classification problems.
In linear case, 
the task is to find the regularization parameter
$\lambda \in \{2^{-20}, 2^{-19}, \ldots, 2^0\}$
that minimizes the LOOCV error. 
In nonlinear case, 
we used Gaussian RBF in the form of 
$\phi_k(\bm x) = \exp(- \gamma \|\bm x - \bm x_k\|^2)$,
where 
$k \in [100]$
were randomly selected from $[n_{\rm old}]$. 
Here,
the task is to select the optimal combination of
$(\lambda, \gamma) \in
\{2^{-15}, 2^{-14}, \ldots, 2^{-5}\}
\times
\{2^{-5}, 2^{-4}, \ldots, 2^5\}
$
that minimizes the LOOCV error.

For further speed-up, we also conducted experiments with two simple tricks. 
In the first trick
we used the lower and the upper bounds of the LOOCV error itself\footnote{
Note that, 
when one already knows that some of the left-instances are correctly classified or not,
the LOOCV error itself can be bounded.
}.
If
the lower bound of one model 
is greater than
the upper bound of another model,
the former model would never be selected as the best model,
meaning that the LOOCV error computation process can be stopped. 
The second simple trick is to 
conduct incremental learning operations
in the increasing order of 
$y_h f(\bm x_h; \bm \beta^*_{\rm old})$.
It is based on a simple observation that 
the class label of an instance whose
$y_h f(\bm x_h; \bm \beta^*_{\rm old})$
value is small tends to be mis-classified.
Note that these two tricks can be used not only for our proposed framework,
but also for other competing approaches.

Tables
\ref{tab:exp-result15}
and
\ref{tab:exp-result16}
show the results without and with the tricks,
respectively.
We see that the computational cost of our proposed framework 
(especially {\bf op2})
are much smaller than competing methods.
It indicates that our bounds in
\eq{eq:bounds-LOOCV}
is tighter
in many cases 
than
the existing bounds for LOOCV error computation.

\section{Conclusions and future works}
\label{sec:conc}
In this paper
we introduced a novel framework for sensitivity analysis of large scale classification problems.
The proposed framework provides a lower and an upper bounds of
a general linear score on the updated classifier
without actually re-optimized it.
The advantage of the proposed framework is that
the computational cost only depends on the sizes of the modified instances,
which is particularly advantageous
in typical sensitivity analysis task
where only relatively small number of instances are updated. 
We discussed three tasks
to which the proposed framework can be applied. 
As a future work,
we plan to apply the proposed framework to stream learning.

\section{Acknowledgement}
For this work, IT was partially supported by JST CREST, MEXT Kakenhi
26106513, and MEXT Kakenhi 26106513.


\appendix

\section{Proofs}
\label{app:proof}
In this section we prove Theorems~\ref{theo:main} and \ref{theo:grad}.
First 
we present 
the following proposition.
%
\begin{prop}
 \label{prop:VI}
 Consider the following general problem:
 \begin{align}
  \label{eq:general.convex.constrained.problem}
  \min_z ~ \phi(z) ~~~ {\rm s.t.} ~ z \in \cZ, 
 \end{align}
 where
 $\phi: \cZ \to \RR$ 
 is a differentiable convex function and $\cZ$ is a convex set. 
 Then a solution $z^*$ is the optimal solution of
 \eq{eq:general.convex.constrained.problem}
 if and only if
 \begin{align*}
  \nabla \phi(z^*)^\top (z^* - z) \le 0 ~~~ \forall ~ z \in \cZ,
 \end{align*}
 where $\nabla \phi(z^*)$ is the gradient vector of $\phi$ at $z = z^*$.
\end{prop}
See,
for example,
Proposition 2.1.2 in \cite{Bertsekas1999}
or
Section 4.2.3 in \cite{Boyd2004}
for the proof of
Proposition~\ref{prop:VI}.

\begin{proof}[Proof of Theorem~\ref{theo:main}]

 From Proposition~\ref{prop:VI}
 and the optimality of 
 $\bm \beta^*_{\rm new}$
 for the problem 
 \eq{eq:prob2}
  \begin{align}
 %
  \label{eq:prf2}
  \!\!\!\!\! \!\!\!
  \bigg(
  \frac{1}{n_{\rm new}}
  \sum_{i \in \cD_{\rm new}}
  \bm \nabla \ell_i(\bm \beta^*_{\rm new})
  +
  \lambda \bm \beta^*_{\rm new}
  \bigg)^\top
  (\bm \beta^*_{\rm new} - \bm \beta^*_{\rm old})
  \le
  0.
 \end{align}
 From the convexity of 
 $\ell_i$, 
 \begin{align}
  \label{eq:prf3}
  &
  \ell_i(\bm \beta^*_{\rm old}) \:
  \ge 
  \ell_i(\bm \beta^*_{\rm new})
  +
  \bm \nabla \ell_i (\bm \beta^*_{\rm new})^\top
  (\bm \beta^*_{\rm old} - \bm \beta^*_{\rm new}),
  \\
  \label{eq:prf4}
  &
  \ell_i(\bm \beta^*_{\rm new}) 
  \ge 
  \ell_i(\bm \beta^*_{\rm old\:}) 
  +
  \bm \nabla \ell_i (\bm \beta^*_{\rm old\:})^\top 
  (\bm \beta^*_{\rm new} - \bm \beta^*_{\rm old}).
 \end{align}

 Using \eq{eq:prf3} and \eq{eq:prf4}, 
  \begin{align}
  \label{eq:prf5}
    \bm \nabla \ell_i (\bm \beta^*_{\rm new})^\top
  (\bm \beta^*_{\rm new} - \bm \beta^*_{\rm old})
  \ge
   \bm \nabla \ell_i (\bm \beta^*_{\rm old\:})^\top 
  (\bm \beta^*_{\rm new} - \bm \beta^*_{\rm old}).
 \end{align}

 By summing up \eq{eq:prf5} for all $i \in \cD_{\rm new}$, 
    \begin{align}
  \label{eq:prf6}
  \sum_{i \in \cD_{\rm new}}  \bm \nabla \ell_i (\bm \beta^*_{\rm new})^\top
  (\bm \beta^*_{\rm new} - \bm \beta^*_{\rm old})
  \ge
  \sum_{i \in \cD_{\rm new}} \bm \nabla \ell_i (\bm \beta^*_{\rm old\:})^\top 
  (\bm \beta^*_{\rm new} - \bm \beta^*_{\rm old}).
 \end{align}
 Substituting \eq{eq:prf6} into  \eq{eq:prf2}, 
  \begin{align}
  \label{eq:prf7}
  \!\!\!\!\! \!\!\!
  \frac{1}{n_{\rm new}}
  \sum_{i \in \cD_{\rm new}}
  \bm \nabla \ell_i(\bm \beta^*_{\rm old})^{\top}
  (\bm \beta^*_{\rm new} - \bm \beta^*_{\rm old})
  +
  \lambda \bm \beta^{*\top}_{\rm new}
  (\bm \beta^*_{\rm new} - \bm \beta^*_{\rm old})
  \le
  0.
 \end{align}
 By completing the square of 
 \eq{eq:prf7}, 
 we have 
\begin{align}
   \label{eq:prf8}
   \Bigg\|
   \bm \beta^*_{\rm new}-\frac{1}{2}
   \Bigg(
   \bm \beta^*_{\rm old}-\frac{\lambda^{-1}}{n_{\rm new}}\sum_{i \in \cD_{\rm new}}
  \bm \nabla \ell_i(\bm \beta^*_{\rm old})
  \Bigg)
  \Bigg\|^{2}
  \le
  \Bigg(
	\frac{1}{2}
   \Bigg\|
   \bm \beta^*_{\rm old}+\frac{\lambda^{-1}}{n_{\rm new}}\sum_{i \in \cD_{\rm new}}
  \bm \nabla \ell_i(\bm \beta^*_{\rm old})
   \Bigg\|
   \Bigg)^{2}.
  \end{align}
 
Furthermore,
noting that
$\bm \beta^*_{\rm old}$ is the optimal solution of
\eq{eq:prob1}, 
 \begin{align}
  \label{eq:prf9}
  \bm \beta^*_{\rm old}+\frac{\lambda^{-1}}{n_{\rm old}}\sum_{i \in \cD_{\rm old}}\nabla \ell_i(\bm \beta^*_{\rm old})=0.
   \end{align}
 Using \eq{eq:prf9} and \eq{eq:delta-s}, 
 \begin{align}
 \frac{\lambda^{-1}}{n_{\rm new}}\sum_{i \in \cD_{\rm new}}
  \bm \nabla \ell_i(\bm \beta^*_{\rm old})
 &
 =\frac{\lambda^{-1}}{n_{\rm new}}
\Bigg(
\sum_{i \in \cD_{\rm old}}\bm \nabla \ell_i(\bm \beta^*_{\rm old})
+
\sum_{i \in \cA}\bm \nabla \ell_i(\bm \beta^*_{\rm old})
-
\sum_{i \in \cR}\bm \nabla \ell_i(\bm \beta^*_{\rm old})
\Bigg)\nonumber\\
&
=\frac{\lambda^{-1}}{n_{\rm new}}
\Bigg(
-\lambda n_{\rm old}\bm \beta^*_{\rm old}+(n_A+n_R) \bm \Delta s
\Bigg)\nonumber\\
&
 \label{eq:prf10}
=-\frac{n_{\rm old}}{n_{\rm new}}
\bm \beta^*_{\rm old}+\frac{(n_A+n_R)\lambda^{-1}}{n_{\rm new}}\bm \Delta s
 \end{align}
 Substituting  \eq{eq:prf10} into \eq{eq:prf8}, 
   \begin{align}
   \label{eq:prf11}
 \Bigg\|
  \bm \beta^*_{\rm new}-
  \Bigg(
  \frac{n_{\rm old} + n_{\rm new}}{2 n_{\rm new}} \bm \beta^*_{\rm old} - \frac{(n_A + n_R)\lambda^{-1}}{2 n_{\rm new}} \bm \Delta s
  \Bigg)
  \Bigg\|^2
  \le
  \Bigg(
  \frac{1}{2}
  \Bigg\|
  \frac{n_A - n_R}{n_{\rm new}} \bm \beta^*_{\rm old}
  +
  \frac{(n_A + n_R) \lambda^{-1}}{n_{\rm new}} \bm \Delta s
  \Bigg\|
  \Bigg)^2
  \end{align}

 Let 
\begin{align}
\label{eq:prf11}
  \bm m =   
  \frac{n_{\rm old} + n_{\rm new}}{2 n_{\rm new}} \bm \beta^*_{\rm old} 
  - \frac{(n_A + n_R)\lambda^{-1}}{2 n_{\rm new}} \bm \Delta s
  ~,~
  r= 
  \frac{1}{2}
  \Bigg\|
  \frac{n_A - n_R}{n_{\rm new}} \bm \beta^*_{\rm old}
  +
  \frac{(n_A + n_R) \lambda^{-1}}{n_{\rm new}} \bm \Delta s
  \Bigg\|.
\end{align}
Then,
 \eq{eq:prf11}
 is compactly written as
\begin{align}
\label{eq:prf12}
 \bm \beta^*_{\rm new} \in \Omega,
 \text{ where }
  \Omega := \{
  \bm \beta
  ~|~
  \|
  \bm \beta -
  \bm m
  \|^2
  \le
  r^2
  \}. 
\end{align}

Eq. \eq{eq:prf12} indicates that the new optimal solution
$\bm \beta^*_{\rm new}$
is within a ball with center $\bm m$ and radius $r$.
Thus,
we have a lower and an upper bounds of a linear score
 $\bm \eta^\top \bm \beta^*_{\rm new}$
 as follows:
  \begin{align}
  \label{eq:LB}
  L(\bm \eta^\top \bm \beta^*_{\rm new})
  &:=
  \min_{\bm \beta \in \Omega}
  ~
  \bm \eta^\top \bm \beta,
  \\
  \label{eq:UB}
  U(\bm \eta^\top \bm \beta^*_{\rm new})
  &:=
  \max_{\bm \beta \in \Omega}
  ~
  \bm \eta^\top \bm \beta.
 \end{align}
 In fact,
 the solutions of \eq{eq:LB} and \eq{eq:UB}
 can be analytically obtained,
 and thus the lower bound 
 $L(\bm \eta^\top \bm \beta^*_{\rm new})$
 and the upper bound
 $U(\bm \eta^\top \bm \beta^*_{\rm new}) $
 can be explicitly obtained 
 by using Lagrange multiplier method.
 Using a Lagrange multiplier
 $\alpha > 0$, 
 the problem
 \eq{eq:LB}
 is rewritten as
 \begin{align*}
  &
  \min_{\bm \beta}~\bm \eta^\top \bm \beta
  ~~~
  {\rm s.t.}
  ~
  \|\bm \beta - \bm m\|^2 \le r^2
  \\
  =
  &
  \min_{\bm \beta}
  \max_{\alpha > 0}
  \big(
  \bm \eta^\top \bm \beta + \alpha (\|\bm \beta - \bm m\|^2 - r^2)
  \big)
  \\
  =
  &
  \max_{\alpha > 0}
  \big(
  - \alpha r^2
  +
  \min_{\bm \beta}
  \big(
  \alpha \|\bm \beta - \bm m\|^2 + \bm \eta^\top \bm \beta
  \big)
  \big)
  \\
  =
  &
  \max_{\alpha > 0}
  ~
  H(\alpha) := 
  \big(
  - \alpha r^2
  - \frac{\|\bm \eta\|^2}{4 \alpha}
  +
  \bm \eta^\top \bm m
  \big),
 \end{align*}
 where
 $\alpha$
 is strictly positive 
 because the constraint
 $\|\bm \beta - \bm m\|^2 \le r^2$
 is strictly active
 at the optimal solution.
 By letting
 $\partial H(\alpha)/\partial \alpha = 0$,
 the optimal
 $\alpha$
 is written as
 \begin{align*}
  \alpha^* := \frac{\|\bm \eta\|}{2 r}
  =
  \arg \max_{\alpha > 0} ~ H(\alpha).
 \end{align*}
 Substituting
 $\alpha^*$
 into
 $H(\alpha)$,
 \begin{align*}
  \bm \eta^\top \bm m - \|\bm \eta\| r = \max_{\alpha > 0}~H(\alpha).
 \end{align*}
Therefore,   
  \begin{align}
  \label{eq:LB2}
  L(\bm \eta^\top \bm \beta^*_{\rm new})
  &=
  \min_{\bm \beta \in \Omega}
  ~
  \bm \eta^\top \bm \beta
  =\bm \eta^{\top}\bm m - \|\bm \eta\|r
 \end{align}
 The upper bound of
 $\bm \eta^\top \bm \beta$
 in
 \eq{eq:UB}
 can be similarly obtained as
   \begin{align}
    \label{eq:UB2}
  U(\bm \eta^\top \bm \beta^*_{\rm new})
  &=
  \max_{\bm \beta \in \Omega}
  ~
  \bm \eta^\top \bm \beta
  =\bm \eta^{\top}\bm m + \|\bm \eta\|r.
 \end{align}
 By substituting
 $\bm m$
 and
 $r$
 in
 \eq{eq:prf11}
 into 
 \eq{eq:LB2}
 and 
 \eq{eq:UB2},
 we have
 \eq{eq:main-LB}
 and 
 \eq{eq:main-UB}. 
\end{proof}

Theorem \ref{theo:grad} can be shown in a similar way as above.
\begin{proof}[Proof of Theorem~\ref{theo:grad}]
 From
 Proposition~\ref{prop:VI}
 and
 the optimality of
 $\bm \beta^{*}_{\rm new}$,
 \begin{align}
  \label{eq:prf36}
  \!\!\!\!\! \!\!\!
  \bigg(
  \frac{1}{n_{\rm new}}
  \sum_{i \in \cD_{\rm new}}
  \bm \nabla \ell_i(\bm \beta^*_{\rm new})
  +
  \lambda \bm \beta^*_{\rm new}
  \bigg)^\top
  (\bm \beta^*_{\rm new} - \hat{\bm \beta}_{\rm new})
  \le
  0.
 \end{align}
 From the convexity of
 $\ell_i$,
 \begin{align}
  \label{eq:prf37}
  \ell_i(\bm \beta^*_{\rm new}) \ge \ell_i(\hat{\bm \beta}_{\rm new}) + \bm \nabla \ell_i(\hat{\bm \beta}_{\rm new})^\top (\bm \beta^*_{\rm new} - \hat{\bm \beta}_{\rm new}).
  \\
  \label{eq:prf38}
  \ell_i(\hat{\bm \beta}_{\rm new}) \ge \ell_i(\bm \beta^*_{\rm new}) + \bm \nabla \ell_i(\bm \beta^*_{\rm new})^\top (\hat{\bm \beta}_{\rm new} - \bm \beta^*_{\rm new}).
 \end{align}
 Using
 \eq{eq:prf37} and \eq{eq:prf38}
 and
 the definition of
 $\bm g(\hat{\bm \beta}_{\rm new})$, 
 the inequality 
 \eq{eq:prf36} 
 can be rewritten as the following condition on 
 the new optimal solution 
 $\bm \beta^*_{\rm new}$: 
 \begin{align}
  \label{eq:prf39}
  \bm \beta^*_{\rm new} \in \hat{\Omega}, 
 \end{align}
 where
 \begin{align*}
  \hat{\Omega}
  :=  \Bigg\{
  \bm \beta
  \Bigg|~
  \Bigg\|
  \bm \beta -
  \Bigg(
  \hat{\bm \beta}_{\rm new} - \frac{\lambda^{-1}}{2} \bm g(\hat{\bm \beta}_{\rm new})
  \Bigg)
  \Bigg\|^2
  \le
  \Bigg(
  \frac{\lambda^{-1}}{2} \| \bm g(\hat{\bm \beta}_{\rm new})\|
  \Bigg)^2
  \Bigg\}.
 \end{align*}
 Then,
 the lower and the upper bounds in
 \eq{eq:grad-bound}
 are obtained by solving the following minimization and maximization problems
 \begin{align}
  \label{eq:prf14}
  L(\bm \eta^\top \bm \beta^*_{\rm new})
  &:=
  \min_{\bm \beta \in \hat{\Omega}}
  ~
  \bm \eta^\top \bm \beta,
  \\
  \label{eq:prf15}
  U(\bm \eta^\top \bm \beta^*_{\rm new})
  &:=
  \max_{\bm \beta \in \hat{\Omega}}
  ~
  \bm \eta^\top \bm \beta.
 \end{align}
 Using the standard Lagrange multiplier method,
 the solutions of
 \eq{eq:prf14}
 and 
 \eq{eq:prf15}
 are analytically obtained as
 \eq{eq:grad-LB}
 and 
 \eq{eq:grad-UB},
 respectively.
\end{proof}


\clearpage
\begin{figure*}[p]
 \begin{center}
  \begin{tabular}{cc}
   \includegraphics[width=0.5\textwidth]{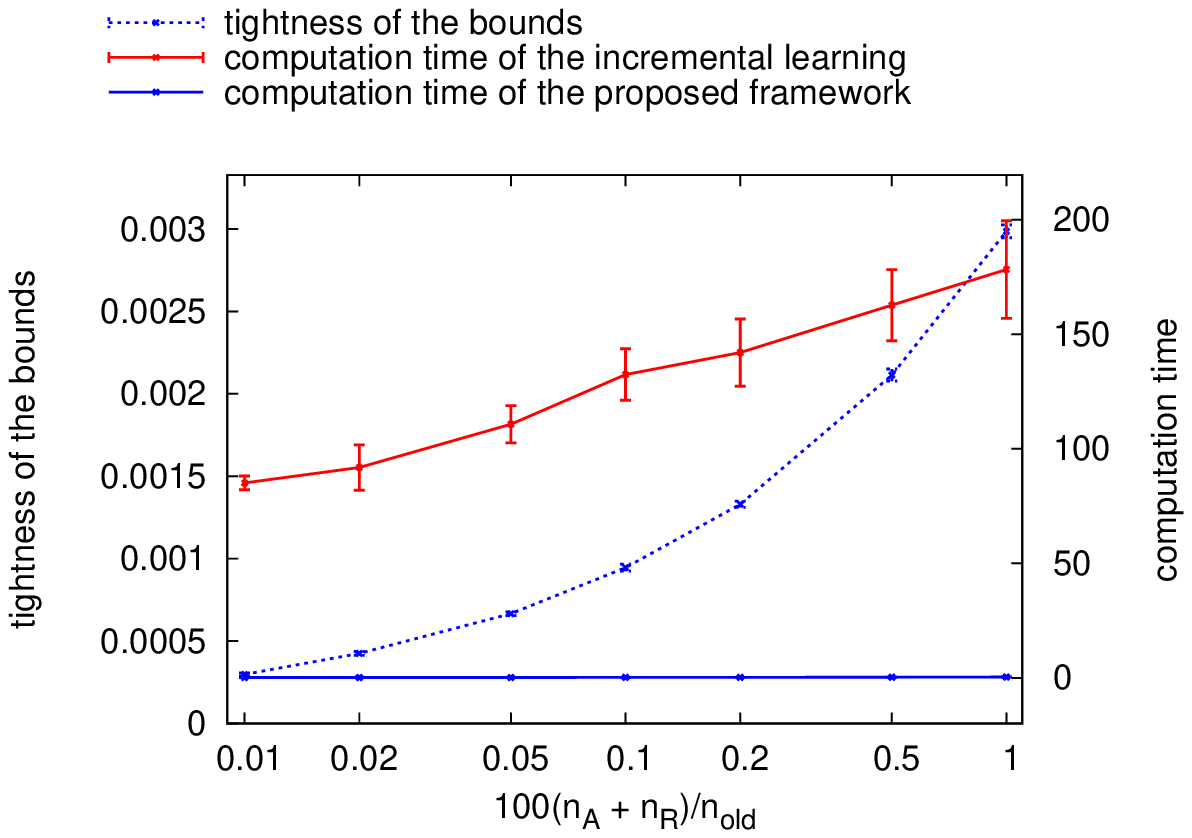} &
   \includegraphics[width=0.5\textwidth]{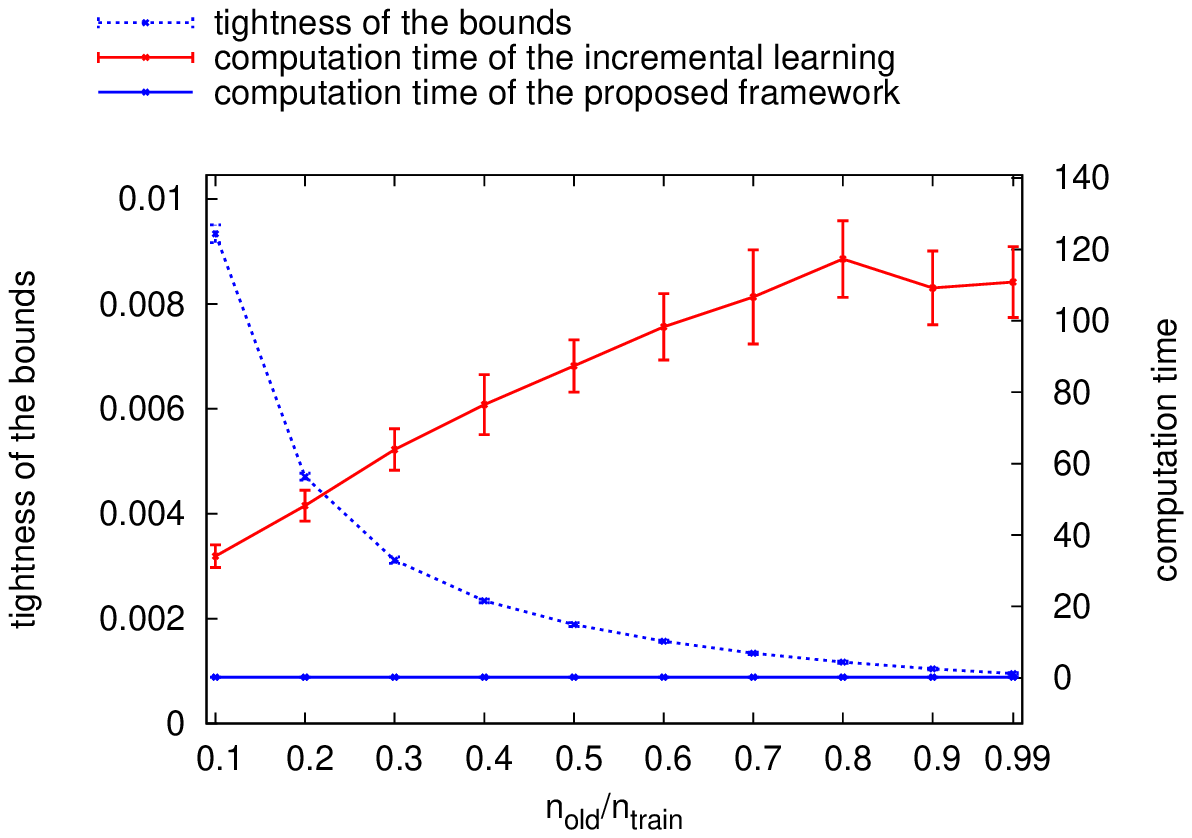} \\
      \includegraphics[width=0.5\textwidth]{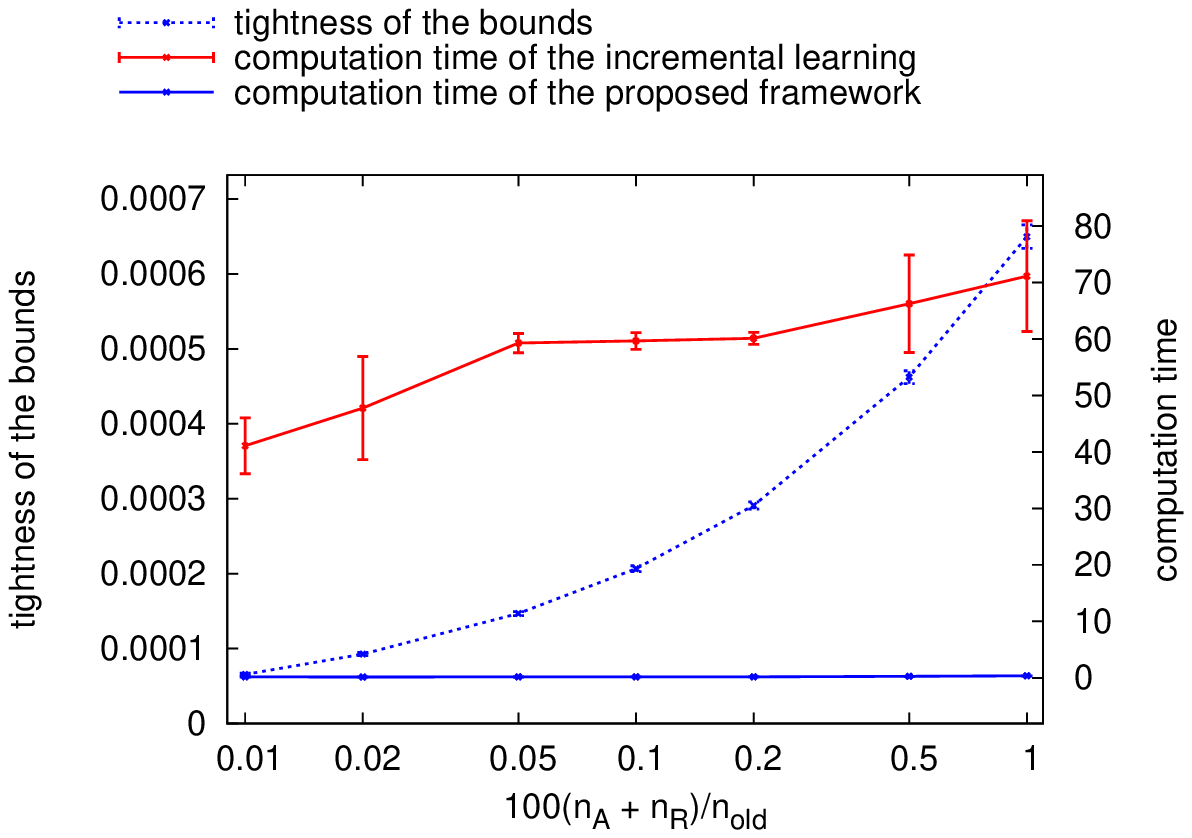} &
   \includegraphics[width=0.5\textwidth]{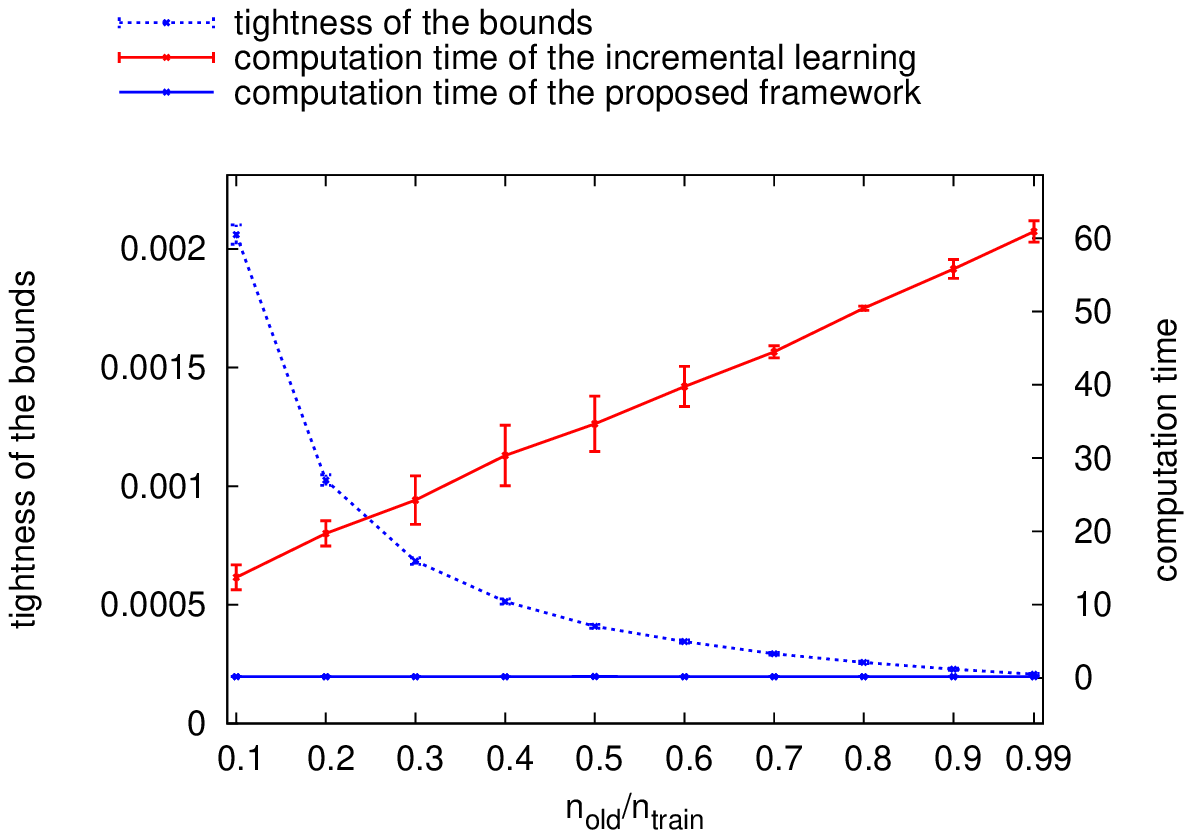} \\
      \includegraphics[width=0.5\textwidth]{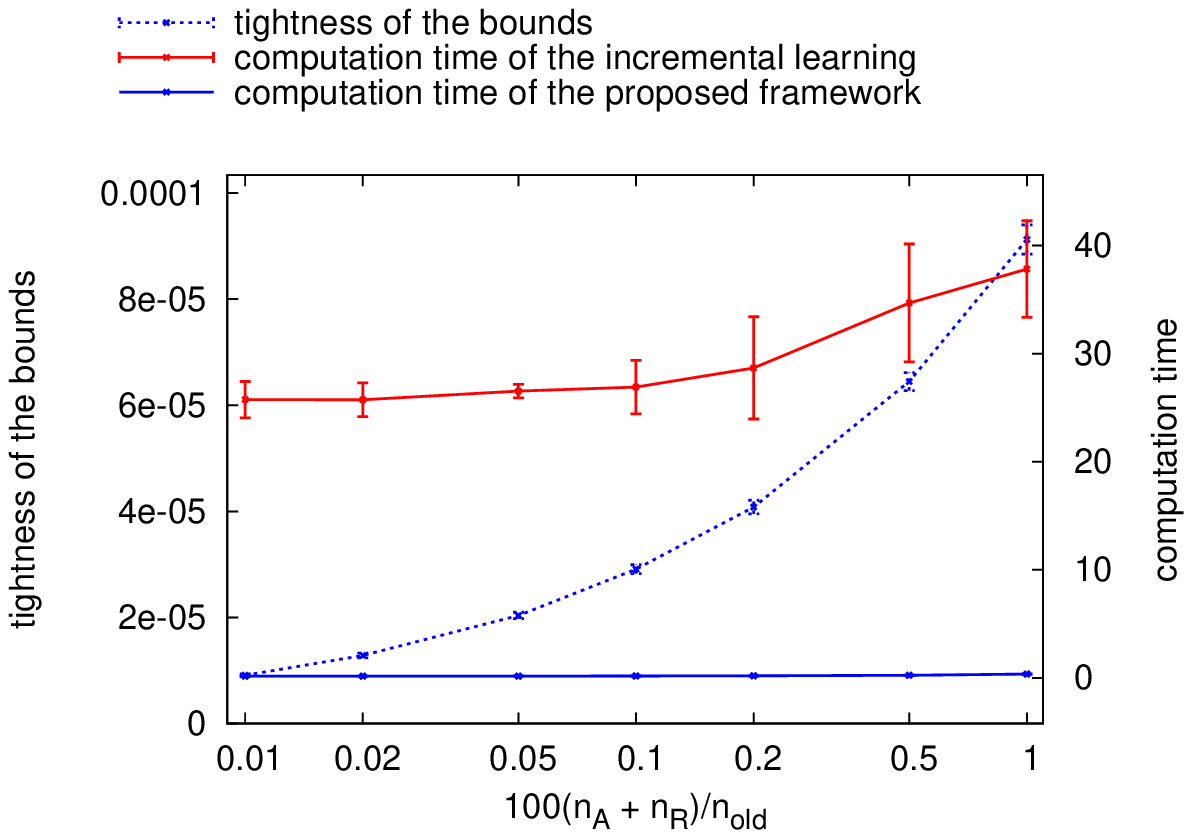} &
   \includegraphics[width=0.5\textwidth]{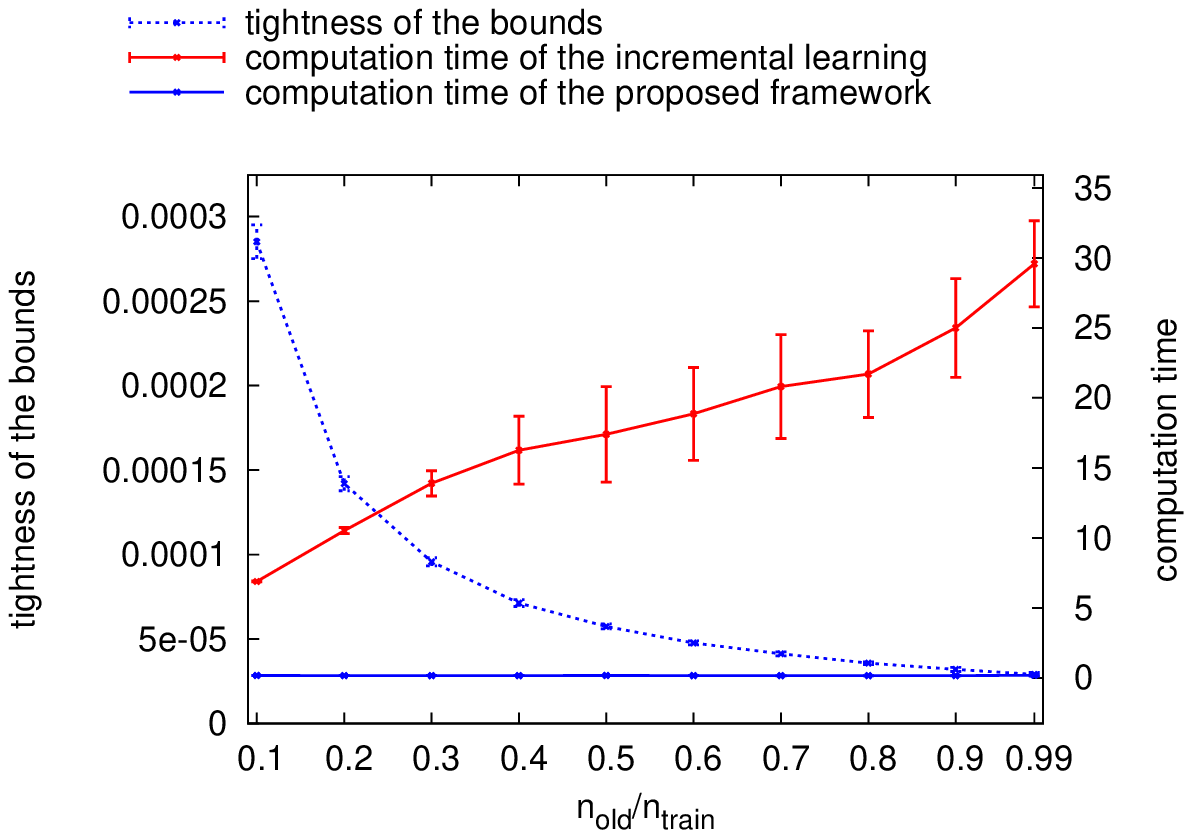} 
  \end{tabular}
  \caption{Results on sensitivity analysis of coefficients for {\tt D8}. The tightness of the bounds and the computation time in seconds are plotted for $\lambda = 0.01$ (top), $0.1$ (middle), and $1$ (bottom).}
  \label{fig:exp-result1}
 \end{center}
\end{figure*}

\begin{figure*}[t]
 \begin{center}
  \begin{tabular}{cc}
   \includegraphics[width=0.5\textwidth]{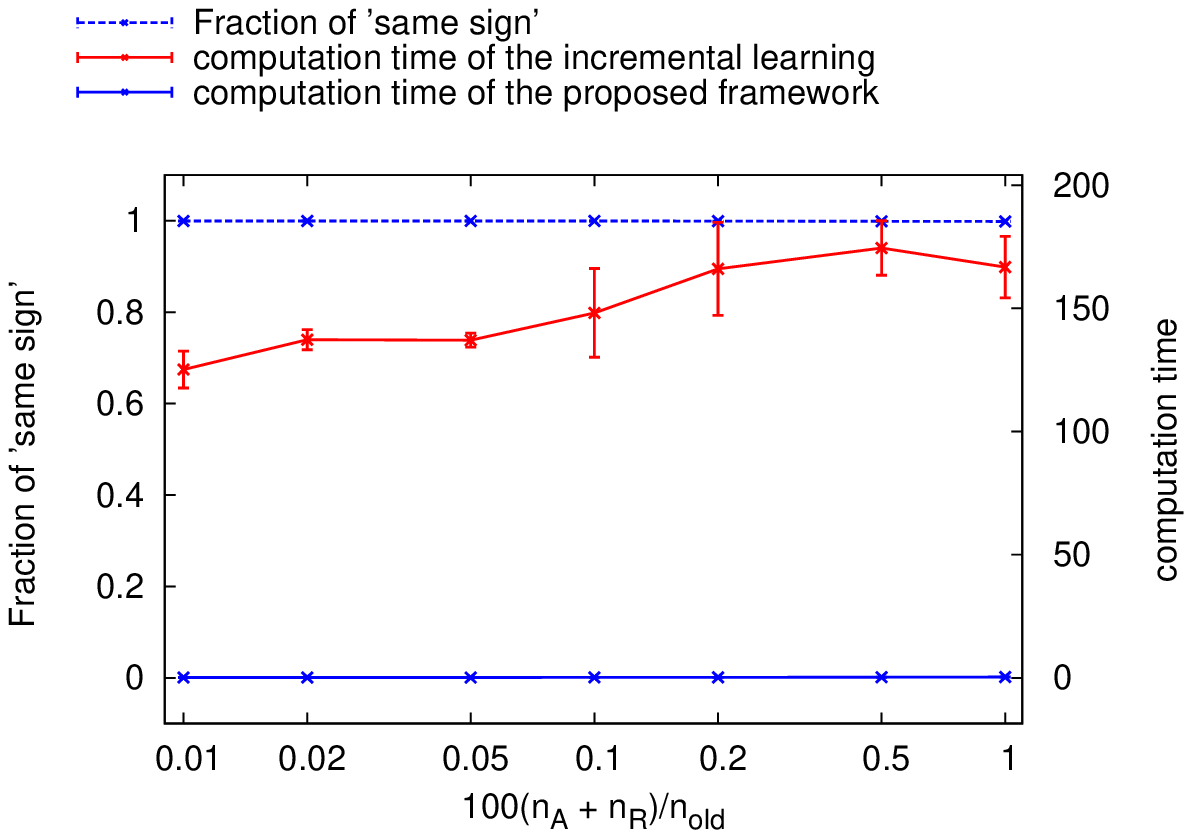} &
   \includegraphics[width=0.5\textwidth]{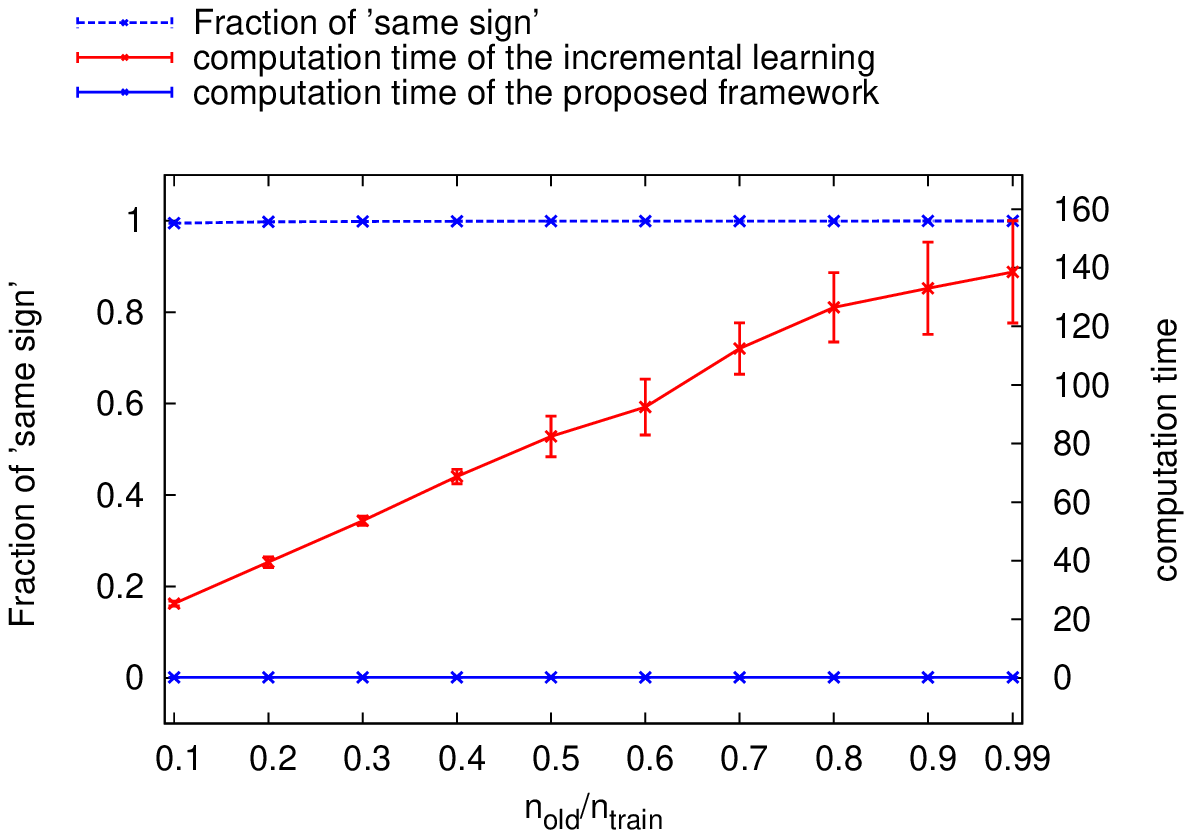} \\
      \includegraphics[width=0.5\textwidth]{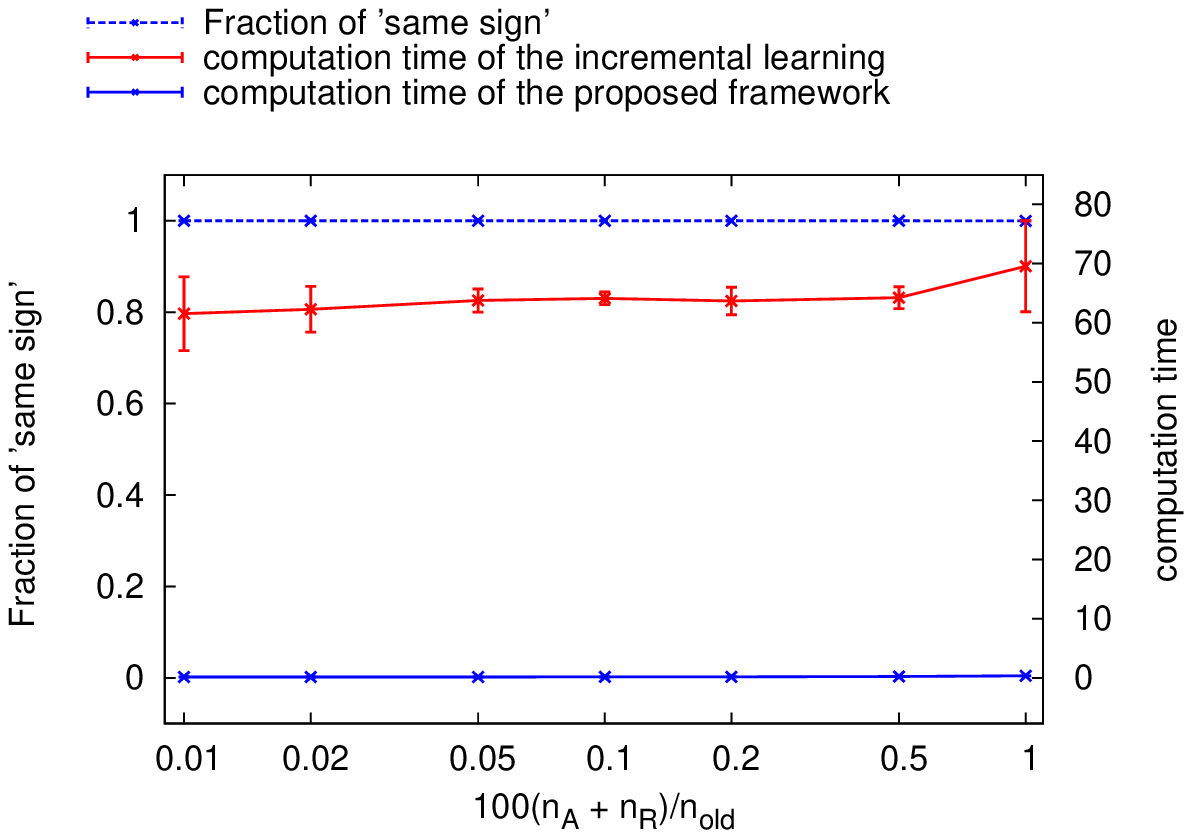} &
   \includegraphics[width=0.5\textwidth]{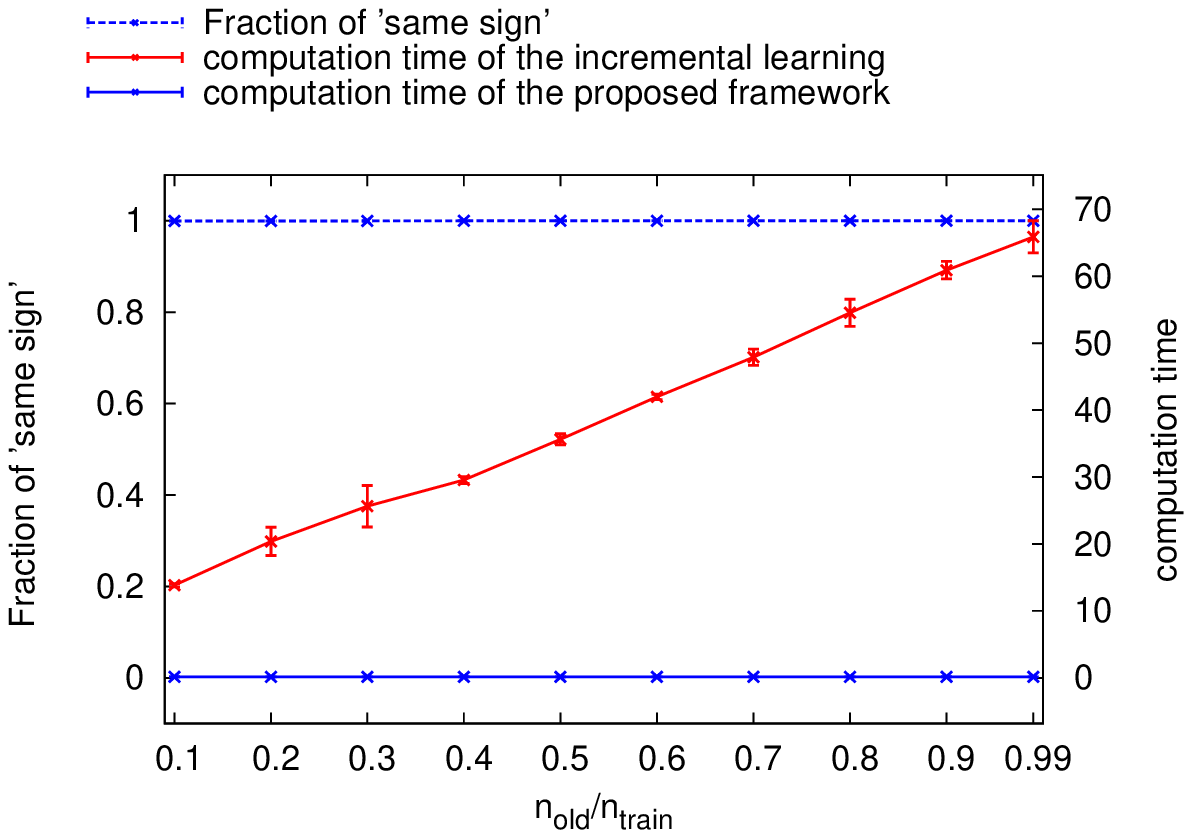} \\
      \includegraphics[width=0.5\textwidth]{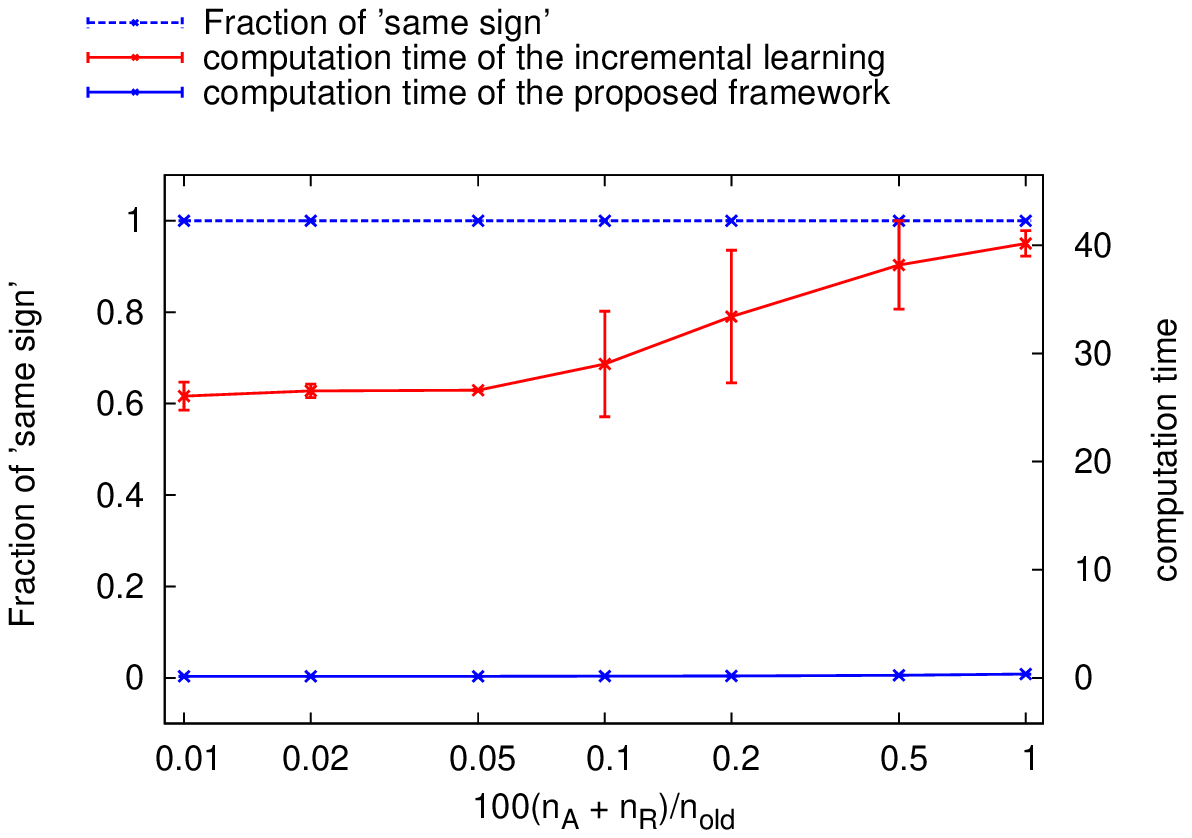} &
   \includegraphics[width=0.5\textwidth]{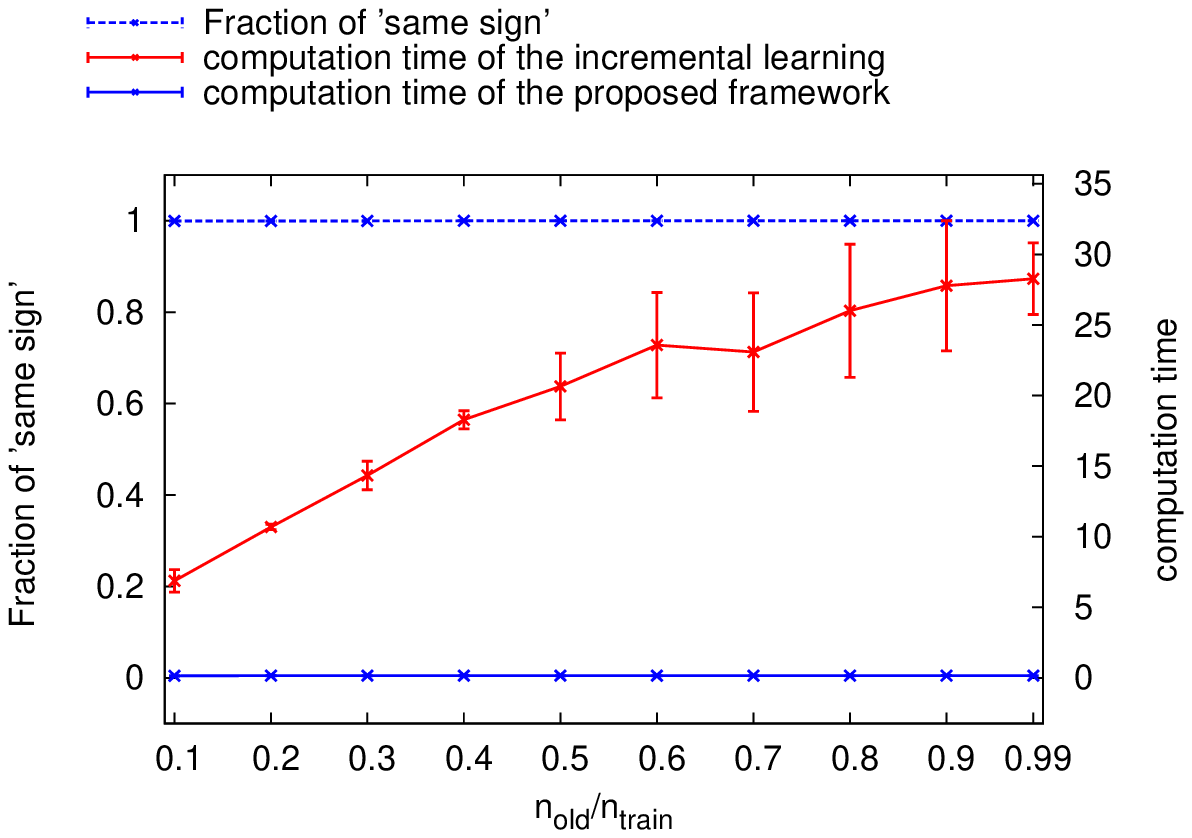} 
  \end{tabular}
 \end{center}
  \caption{Results on sensitivity analysis of class labels for {\tt D8}. The fraction of the test instances whose lower and upper bounds of the decision score have same signs, and the computation time in seconds are plotted for $\lambda = 0.01$ (top), $0.1$ (middle), and $1$ (bottom).}
  \label{fig:exp-result2}
\end{figure*}

\clearpage

\begin{table}[p]
\begin{center}
\begin{scriptsize}
\begin{tabular}{|c|c||c|c||c|c||c|c|}
\hline
\multicolumn{2}{|c||}{}&\multicolumn{6}{|c|}{$(n_{\rm A}+n_{\rm R})/n_{\rm old}$}\\
\cline{3-8}
\multicolumn{2}{|c||}{}&\multicolumn{2}{|c||}{$0.01\%$}&\multicolumn{2}{|c||}{$0.1\%$}&\multicolumn{2}{|c|}{$1\%$}\\
\cline{3-8}
\multicolumn{2}{|c||}{}&Incremental&proposed&Incremental&proposed&Incremental&proposed\\
\hline
\hline
 D5 & tightness& N.A. & 5.68e-03& N.A. & 1.94e-02& N.A. & 6.63e-02\\
 & & &($\pm$3.29e-03)&&($\pm$5.69e-03)&&($\pm$1.72e-02) \\
\cline{2-8}
 & time [sec]& 4.57e-02 & 4.47e-06 & 5.41e-02 & 2.26e-05 & 6.16e-02 & 1.95e-04 \\ 
 & &($\pm$6.38e-03) &($\pm$4.99e-07) &($\pm$2.37e-03) &($\pm$4.96e-07) &($\pm$5.32e-03) &($\pm$7.45e-07)\\ 
\hline
\hline
 D6 & tightness& N.A. & 1.31e-03& N.A. & 5.08e-03& N.A. & 1.56e-02\\
 & & &($\pm$8.35e-04)&&($\pm$9.33e-04)&&($\pm$2.92e-03) \\
\cline{2-8}
 & time [sec]& 8.42e-02 & 6.37e-06 & 8.58e-02 & 3.21e-05 & 9.97e-02 & 2.84e-04 \\ 
 & &($\pm$1.87e-02) &($\pm$2.50e-06) &($\pm$8.23e-03) &($\pm$1.38e-06) &($\pm$1.73e-02) &($\pm$7.55e-06)\\ 
\hline
\hline
 D7 & tightness& N.A. & 2.00e-03& N.A. & 6.83e-03& N.A. & 2.27e-02\\
 & & &($\pm$7.78e-04)&&($\pm$2.33e-03)&&($\pm$9.20e-03) \\
\cline{2-8}
 & time [sec]& 1.30e-01 & 6.47e-06 & 1.65e-01 & 3.03e-05 & 1.90e-01 & 2.38e-04 \\ 
 & &($\pm$2.88e-02) &($\pm$1.18e-06) &($\pm$2.13e-02) &($\pm$3.34e-06) &($\pm$2.36e-02) &($\pm$2.31e-05)\\ 
\hline
\hline
 D8 & tightness& N.A. & 2.98e-04& N.A. & 9.45e-04& N.A. & 2.98e-03\\
 & & &($\pm$8.96e-06)&&($\pm$2.11e-05)&&($\pm$3.89e-05) \\
\cline{2-8}
 & time [sec]& 8.51e+01 & 1.53e-01 & 1.32e+02 & 1.82e-01 & 1.78e+02 & 3.59e-01 \\ 
 & &($\pm$3.00e+00) &($\pm$8.24e-03) &($\pm$1.13e+01) &($\pm$1.17e-02) &($\pm$2.13e+01) &($\pm$1.63e-02)\\ 
\hline
\end{tabular}
\end{scriptsize}
\caption{Results on sensitivity analysis of coefficients for various values of $(n_{\rm A}+n_{\rm R})/n_{\rm old}$. The tightness of the bounds and the computation time in seconds are listed ($\lambda = 0.01$).}
\label{tab:exp-result3}
\end{center}
\end{table}

\begin{table}[p]
\begin{center}
\begin{scriptsize}
\begin{tabular}{|c|c||c|c||c|c||c|c|}
\hline
\multicolumn{2}{|c||}{}&\multicolumn{6}{|c|}{$(n_{\rm A}+n_{\rm R})/n_{\rm old}$}\\
\cline{3-8}
\multicolumn{2}{|c||}{}&\multicolumn{2}{|c||}{$0.01\%$}&\multicolumn{2}{|c||}{$0.1\%$}&\multicolumn{2}{|c|}{$1\%$}\\
\cline{3-8}
\multicolumn{2}{|c||}{}&Incremental&proposed&Incremental&proposed&Incremental&proposed\\
\hline
\hline
 D5 & tightness& N.A. & 7.55e-04& N.A. & 2.27e-03& N.A. & 6.49e-03\\
 & & &($\pm$3.74e-04)&&($\pm$1.26e-03)&&($\pm$1.83e-03) \\
\cline{2-8}
 & time [sec]& 3.03e-02 & 4.13e-06 & 4.29e-02 & 2.21e-05 & 4.45e-02 & 1.94e-04 \\ 
 & &($\pm$3.24e-03) &($\pm$4.27e-07) &($\pm$4.89e-03) &($\pm$6.29e-07) &($\pm$1.85e-03) &($\pm$8.23e-07)\\ 
\hline
\hline
 D6 & tightness& N.A. & 1.86e-04& N.A. & 6.54e-04& N.A. & 2.09e-03\\
 & & &($\pm$3.86e-05)&&($\pm$1.10e-04)&&($\pm$3.62e-04) \\
\cline{2-8}
 & time [sec]& 3.17e-02 & 4.63e-06 & 6.49e-02 & 3.09e-05 & 6.60e-02 & 2.80e-04 \\ 
 & &($\pm$7.71e-03) &($\pm$8.36e-07) &($\pm$6.65e-03) &($\pm$1.45e-06) &($\pm$7.14e-03) &($\pm$6.50e-06)\\ 
\hline
\hline
 D7 & tightness& N.A. & 2.14e-04& N.A. & 6.60e-04& N.A. & 2.34e-03\\
 & & &($\pm$7.57e-05)&&($\pm$2.80e-04)&&($\pm$9.89e-04) \\
\cline{2-8}
 & time [sec]& 8.41e-02 & 6.30e-06 & 1.02e-01 & 3.08e-05 & 1.13e-01 & 2.31e-04 \\ 
 & &($\pm$1.47e-02) &($\pm$9.36e-07) &($\pm$1.45e-02) &($\pm$4.94e-06) &($\pm$1.42e-02) &($\pm$1.43e-05)\\ 
\hline
\hline
 D8 & tightness& N.A. & 6.56e-05& N.A. & 2.07e-04& N.A. & 6.50e-04\\
 & & &($\pm$1.65e-06)&&($\pm$3.94e-06)&&($\pm$1.57e-05) \\
\cline{2-8}
 & time [sec]& 4.11e+01 & 1.61e-01 & 5.96e+01 & 1.72e-01 & 7.11e+01 & 3.53e-01 \\ 
 & &($\pm$4.95e+00) &($\pm$1.03e-02) &($\pm$1.50e+00) &($\pm$8.33e-03) &($\pm$9.80e+00) &($\pm$1.15e-02)\\ 
\hline
\end{tabular}
\end{scriptsize}
\caption{Results on sensitivity analysis of coefficients for various values of $(n_{\rm A}+n_{\rm R})/n_{\rm old}$. The tightness of the bounds and the computation time in seconds are listed ($\lambda = 0.1$).}
\label{tab:exp-result4}
\end{center}
\end{table}

\begin{table}[p]
\begin{center}
\begin{scriptsize}
\begin{tabular}{|c|c||c|c||c|c||c|c|}
\hline
\multicolumn{2}{|c||}{}&\multicolumn{6}{|c|}{$(n_{\rm A}+n_{\rm R})/n_{\rm old}$}\\
\cline{3-8}
\multicolumn{2}{|c||}{}&\multicolumn{2}{|c||}{$0.01\%$}&\multicolumn{2}{|c||}{$0.1\%$}&\multicolumn{2}{|c|}{$1\%$}\\
\cline{3-8}
\multicolumn{2}{|c||}{}&Incremental&proposed&Incremental&proposed&Incremental&proposed\\
\hline
\hline 
 D5 & tightness& N.A. & 7.49e-05& N.A. & 2.56e-04& N.A. & 7.47e-04\\
 & & &($\pm$1.61e-05)&&($\pm$6.87e-05)&&($\pm$1.85e-04) \\
\cline{2-8}
 & time [sec]& 1.45e-02 & 4.10e-06 & 1.89e-02 & 2.08e-05 & 2.26e-02 & 1.88e-04 \\ 
 & &($\pm$1.55e-03) &($\pm$9.07e-07) &($\pm$3.25e-03) &($\pm$2.32e-06) &($\pm$8.96e-04) &($\pm$3.75e-06)\\ 
\hline
\hline
 D6 & tightness& N.A. & 2.09e-05& N.A. & 8.19e-05& N.A. & 2.50e-04\\
 & & &($\pm$3.84e-06)&&($\pm$1.53e-05)&&($\pm$4.33e-05) \\
\cline{2-8}
 & time [sec]& 2.12e-02 & 4.83e-06 & 2.23e-02 & 3.02e-05 & 2.10e-02 & 2.70e-04 \\ 
 & &($\pm$1.49e-03) &($\pm$1.29e-06) &($\pm$8.73e-04) &($\pm$9.45e-07) &($\pm$3.08e-03) &($\pm$5.53e-06)\\ 
\hline
\hline
 D7 & tightness& N.A. & 1.93e-05& N.A. & 8.59e-05& N.A. & 2.21e-04\\
 & & &($\pm$7.66e-06)&&($\pm$3.95e-05)&&($\pm$7.66e-05) \\
\cline{2-8}
 & time [sec]& 1.95e-02 & 4.03e-06 & 2.50e-02 & 2.39e-05 & 2.86e-02 & 2.11e-04 \\ 
 & &($\pm$7.41e-04) &($\pm$4.82e-07) &($\pm$5.21e-03) &($\pm$3.40e-06) &($\pm$3.05e-03) &($\pm$4.99e-06)\\ 
\hline
\hline
 D8 & tightness& N.A. & 9.10e-06& N.A. & 2.91e-05& N.A. & 9.12e-05\\
 & & &($\pm$2.71e-07)&&($\pm$8.31e-07)&&($\pm$2.75e-06) \\
\cline{2-8}
 & time [sec]& 2.58e+01 & 1.59e-01 & 2.69e+01 & 1.71e-01 & 3.78e+01 & 3.48e-01 \\ 
 & &($\pm$1.68e+00) &($\pm$1.04e-02) &($\pm$2.47e+00) &($\pm$5.79e-03) &($\pm$4.46e+00) &($\pm$1.43e-02)\\ 
\hline
\end{tabular}
\end{scriptsize}
\caption{Results on sensitivity analysis of coefficients for various values of $(n_{\rm A}+n_{\rm R})/n_{\rm old}$. The tightness of the bounds and the computation time in seconds are listed ($\lambda = 1$).}
\label{tab:exp-result5}
\end{center}
\end{table}

\begin{table}[p]
\begin{center}
\begin{scriptsize}
\begin{tabular}{|c|c||c|c||c|c||c|c|}
\hline
\multicolumn{2}{|c||}{}&\multicolumn{6}{|c|}{$n_{\rm old}/n_{\rm train}$}\\
\cline{3-8}
\multicolumn{2}{|c||}{}&\multicolumn{2}{|c||}{$10\%$}&\multicolumn{2}{|c||}{$50\%$}&\multicolumn{2}{|c|}{$99\%$}\\
\cline{3-8}
\multicolumn{2}{|c||}{}&Incremental&proposed&Incremental&proposed&Incremental&proposed\\
\hline
\hline
 D5 & tightness& N.A. & 1.92e-01 & N.A. & 3.41e-02 & N.A. & 2.02e-02 \\
  & & & ($\pm$5.96e-02)& & ($\pm$1.29e-02)& &($\pm$6.57e-03) \\
\cline{2-8}
 & time [sec]& 6.88e-03 & 2.14e-05& 4.12e-02 & 2.41e-05& 7.58e-02& 2.64e-05\\ 
 & &($\pm$1.77e-04) &($\pm$6.05e-07) &($\pm$1.01e-02) &($\pm$1.06e-06) &($\pm$1.14e-02) &($\pm$1.30e-06)\\ 
\hline
\hline
 D6 & tightness& N.A. & 5.07e-02 & N.A. & 8.81e-03 & N.A. & 5.13e-03 \\
  & & & ($\pm$1.20e-02)& & ($\pm$1.84e-03)& &($\pm$1.25e-03) \\
\cline{2-8}
 & time [sec]& 5.27e-03 & 2.89e-05& 5.24e-02 & 3.36e-05& 1.01e-01& 3.49e-05\\ 
 & &($\pm$3.00e-04) &($\pm$5.73e-07) &($\pm$1.32e-02) &($\pm$1.28e-06) &($\pm$1.63e-02) &($\pm$1.36e-06)\\ 
\hline
\hline
 D7 & tightness& N.A. & 6.55e-02 & N.A. & 1.10e-02 & N.A. & 6.81e-03 \\
  & & & ($\pm$2.83e-02)& & ($\pm$4.27e-03)& &($\pm$3.09e-03) \\
\cline{2-8}
 & time [sec]& 6.96e-03 & 2.50e-05& 1.12e-01 & 3.06e-05& 1.80e-01& 3.05e-05\\ 
 & &($\pm$5.53e-04) &($\pm$5.46e-06) &($\pm$1.64e-02) &($\pm$1.76e-06) &($\pm$2.87e-02) &($\pm$2.22e-06)\\ 
\hline
\hline
 D8 & tightness& N.A. & 9.34e-03 & N.A. & 1.57e-03 & N.A. & 9.54e-04 \\
  & & & ($\pm$1.68e-04)& & ($\pm$1.95e-05)& &($\pm$1.73e-05) \\
\cline{2-8}
 & time [sec]& 3.41e+01 & 1.63e-01& 9.83e+01 & 1.77e-01& 1.11e+02& 1.60e-01\\ 
 & &($\pm$3.16e+00) &($\pm$8.33e-03) &($\pm$9.30e+00) &($\pm$1.11e-02) &($\pm$9.91e+00) &($\pm$5.66e-03)\\ 
\hline
\end{tabular}
\end{scriptsize}
\caption{Results on sensitivity analysis of coefficients for various values of $n_{\rm old}/n_{\rm train}$. The tightness of the bounds and the computation time in seconds are listed ($\lambda = 0.01$).}
\label{tab:exp-result6}
\end{center}
\end{table}

\begin{table}[p]
\begin{center}
\begin{scriptsize}
\begin{tabular}{|c|c||c|c||c|c||c|c|}
\hline
\multicolumn{2}{|c||}{}&\multicolumn{6}{|c|}{$n_{\rm old}/n_{\rm train}$}\\
\cline{3-8}
\multicolumn{2}{|c||}{}&\multicolumn{2}{|c||}{$10\%$}&\multicolumn{2}{|c||}{$50\%$}&\multicolumn{2}{|c|}{$99\%$}\\
\cline{3-8}
\multicolumn{2}{|c||}{}&Incremental&proposed&Incremental&proposed&Incremental&proposed\\
\hline
\hline
 D5 & tightness& N.A. & 2.31e-02 & N.A. & 3.41e-03 & N.A. & 2.25e-03 \\
  & & & ($\pm$6.87e-03)& & ($\pm$1.03e-03)& &($\pm$8.49e-04) \\
\cline{2-8}
 & time [sec]& 4.16e-03 & 2.17e-05& 2.82e-02 & 2.54e-05& 5.86e-02& 2.81e-05\\ 
 & &($\pm$3.89e-04) &($\pm$6.50e-07) &($\pm$7.70e-03) &($\pm$1.26e-06) &($\pm$1.48e-02) &($\pm$1.54e-06)\\ 
\hline
\hline
 D6 & tightness& N.A. & 6.72e-03 & N.A. & 1.08e-03 & N.A. & 6.60e-04 \\
  & & & ($\pm$1.49e-03)& & ($\pm$2.18e-04)& &($\pm$1.06e-04) \\
\cline{2-8}
 & time [sec]& 3.69e-03 & 2.92e-05& 4.35e-02 & 3.54e-05& 7.67e-02& 3.74e-05\\ 
 & &($\pm$4.51e-04) &($\pm$4.53e-07) &($\pm$1.08e-02) &($\pm$1.69e-06) &($\pm$1.16e-02) &($\pm$1.91e-06)\\ 
\hline
\hline
 D7 & tightness& N.A. & 7.20e-03 & N.A. & 1.23e-03 & N.A. & 7.67e-04 \\
  & & & ($\pm$3.06e-03)& & ($\pm$5.68e-04)& &($\pm$2.88e-04) \\
\cline{2-8}
 & time [sec]& 4.56e-03 & 2.37e-05& 6.21e-02 & 3.14e-05& 1.03e-01& 3.10e-05\\ 
 & &($\pm$5.42e-04) &($\pm$6.29e-07) &($\pm$9.25e-03) &($\pm$2.50e-06) &($\pm$7.35e-03) &($\pm$3.28e-06)\\ 
\hline
\hline
 D8 & tightness& N.A. & 2.06e-03 & N.A. & 3.45e-04 & N.A. & 2.08e-04 \\
  & & & ($\pm$4.06e-05)& & ($\pm$6.89e-06)& &($\pm$3.65e-06) \\
\cline{2-8}
 & time [sec]& 1.37e+01 & 1.63e-01& 3.98e+01 & 1.65e-01& 6.09e+01& 1.70e-01\\ 
 & &($\pm$1.70e+00) &($\pm$9.41e-03) &($\pm$2.75e+00) &($\pm$5.46e-03) &($\pm$1.45e+00) &($\pm$6.76e-03)\\ 
\hline
\end{tabular}
\end{scriptsize}
\caption{Results on sensitivity analysis of coefficients for various values of $n_{\rm old}/n_{\rm train}$. The tightness of the bounds and the computation time in seconds are listed ($\lambda = 0.1$).}
\label{tab:exp-result7}
\end{center}
\end{table}

\begin{table}[p]
\begin{center}
\begin{scriptsize}
\begin{tabular}{|c|c||c|c||c|c||c|c|}
\hline
\multicolumn{2}{|c||}{}&\multicolumn{6}{|c|}{$n_{\rm old}/n_{\rm train}$}\\
\cline{3-8}
\multicolumn{2}{|c||}{}&\multicolumn{2}{|c||}{$10\%$}&\multicolumn{2}{|c||}{$50\%$}&\multicolumn{2}{|c|}{$99\%$}\\
\cline{3-8}
\multicolumn{2}{|c||}{}&Incremental&proposed&Incremental&proposed&Incremental&proposed\\
\hline
\hline
 D5 & tightness& N.A. & 2.48e-03 & N.A. & 4.22e-04 & N.A. & 2.61e-04 \\
  & & & ($\pm$7.00e-04)& & ($\pm$1.17e-04)& &($\pm$7.27e-05) \\
\cline{2-8}
 & time [sec]& 2.42e-03 & 2.05e-05& 1.50e-02 & 2.51e-05& 3.01e-02& 2.79e-05\\ 
 & &($\pm$2.16e-04) &($\pm$6.71e-07) &($\pm$2.38e-03) &($\pm$1.51e-06) &($\pm$5.83e-03) &($\pm$2.31e-06)\\ 
\hline
\hline
 D6 & tightness& N.A. & 7.67e-04 & N.A. & 1.25e-04 & N.A. & 8.32e-05 \\
  & & & ($\pm$1.42e-04)& & ($\pm$2.32e-05)& &($\pm$1.77e-05) \\
\cline{2-8}
 & time [sec]& 2.66e-03 & 2.92e-05& 2.18e-02 & 3.52e-05& 3.92e-02& 3.62e-05\\ 
 & &($\pm$2.74e-04) &($\pm$2.48e-06) &($\pm$4.72e-03) &($\pm$2.26e-06) &($\pm$6.66e-03) &($\pm$3.42e-06)\\ 
\hline
\hline
 D7 & tightness& N.A. & 7.40e-04 & N.A. & 1.27e-04 & N.A. & 7.09e-05 \\
  & & & ($\pm$2.81e-04)& & ($\pm$4.66e-05)& &($\pm$2.39e-05) \\
\cline{2-8}
 & time [sec]& 3.27e-03 & 2.25e-05& 2.66e-02 & 2.90e-05& 3.71e-02& 2.97e-05\\ 
 & &($\pm$6.01e-04) &($\pm$7.19e-07) &($\pm$7.92e-03) &($\pm$1.60e-06) &($\pm$8.81e-03) &($\pm$2.27e-06)\\ 
\hline
\hline
 D8 & tightness& N.A. & 2.85e-04 & N.A. & 4.77e-05 & N.A. & 2.91e-05 \\
  & & & ($\pm$9.97e-06)& & ($\pm$1.18e-06)& &($\pm$1.06e-06) \\
\cline{2-8}
 & time [sec]& 6.89e+00 & 1.64e-01& 1.89e+01 & 1.59e-01& 2.96e+01& 1.80e-01\\ 
 & &($\pm$2.93e-02) &($\pm$8.92e-03) &($\pm$3.32e+00) &($\pm$6.00e-03) &($\pm$3.07e+00) &($\pm$2.07e-02)\\ 
\hline
\end{tabular}
\end{scriptsize}
\caption{Results on sensitivity analysis of coefficients for various values of $n_{\rm old}/n_{\rm train}$. The tightness of the bounds and the computation time in seconds are listed ($\lambda = 1$).}
\label{tab:exp-result8}
\end{center}
\end{table}

\begin{table}[p]
\begin{center}
\begin{scriptsize}
\begin{tabular}{|c|c||c|c||c|c||c|c|}
\hline
\multicolumn{2}{|c||}{}&\multicolumn{6}{|c|}{$(n_{\rm A}+n_{\rm R})/n_{\rm old}$}\\
\cline{3-8}
\multicolumn{2}{|c||}{}&\multicolumn{2}{|c||}{$0.01\%$}&\multicolumn{2}{|c||}{$0.1\%$}&\multicolumn{2}{|c|}{$1\%$}\\
\cline{3-8}
\multicolumn{2}{|c||}{}&Incremental&proposed&Incremental&proposed&Incremental&proposed\\
\hline
\hline 
 D5 & fraction of ``same sign"& N.A. & 9.96345e-01& N.A. & 9.88742e-01& N.A. & 9.65412e-01\\
 & & &($\pm$1.68e-03)&&($\pm$4.33e-03)&&($\pm$1.01e-02) \\
\cline{2-8}
 & time [sec]& 6.80e-02 & 4.15e-04 & 7.89e-02 & 4.36e-04 & 8.82e-02 & 6.38e-04 \\ 
 & &($\pm$1.09e-02) &($\pm$1.90e-05) &($\pm$1.78e-02) &($\pm$1.04e-05) &($\pm$1.98e-02) &($\pm$3.84e-05)\\ 
\hline
\hline
 D6 & fraction of ``same sign"& N.A. & 1.00000e+00& N.A. & 1.00000e+00& N.A. & 1.00000e+00\\
 & & &($\pm$0.00e+00)&&($\pm$0.00e+00)&&($\pm$0.00e+00) \\
\cline{2-8}
 & time [sec]& 1.13e-01 & 2.80e-03 & 1.10e-01 & 2.86e-03 & 1.37e-01 & 3.14e-03 \\ 
 & &($\pm$2.13e-02) &($\pm$1.50e-04) &($\pm$1.38e-02) &($\pm$1.63e-04) &($\pm$1.37e-02) &($\pm$1.20e-04)\\ 
\hline
\hline
 D7 & fraction of ``same sign"& N.A. & 9.99728e-01& N.A. & 9.99354e-01& N.A. & 9.98612e-01\\
 & & &($\pm$5.51e-05)&&($\pm$1.71e-04)&&($\pm$4.63e-04) \\
\cline{2-8}
 & time [sec]& 1.40e-01 & 5.30e-03 & 1.55e-01 & 5.15e-03 & 1.96e-01 & 5.76e-03 \\ 
 & &($\pm$3.26e-02) &($\pm$4.20e-04) &($\pm$3.42e-02) &($\pm$1.31e-04) &($\pm$2.45e-02) &($\pm$4.57e-04)\\ 
\hline
\hline
 D8 & fraction of ``same sign"& N.A. & 9.99869e-01& N.A. & 9.99583e-01& N.A. & 9.98672e-01\\
 & & &($\pm$7.49e-06)&&($\pm$1.76e-05)&&($\pm$9.61e-05) \\
\cline{2-8}
 & time [sec]& 1.25e+02 & 1.40e-01 & 1.48e+02 & 1.67e-01 & 1.67e+02 & 3.49e-01 \\ 
 & &($\pm$7.47e+00) &($\pm$9.27e-03) &($\pm$1.80e+01) &($\pm$8.01e-03) &($\pm$1.25e+01) &($\pm$2.14e-02)\\ 
\hline
\end{tabular}
\end{scriptsize}
\caption{Results on sensitivity analysis on class labels for various values of $(n_{\rm A}+n_{\rm R})/n_{\rm old}$. The fraction of the test instances whose lower and upper bounds of the decision score have same signs, and the computation time in seconds are listed ($\lambda = 0.01$).}
\label{tab:exp-result9}
\end{center}
\end{table}

\begin{table}[p]
\begin{center}
\begin{scriptsize}
\begin{tabular}{|c|c||c|c||c|c||c|c|}
\hline
\multicolumn{2}{|c||}{}&\multicolumn{6}{|c|}{$(n_{\rm A}+n_{\rm R})/n_{\rm old}$}\\
\cline{3-8}
\multicolumn{2}{|c||}{}&\multicolumn{2}{|c||}{$0.01\%$}&\multicolumn{2}{|c||}{$0.1\%$}&\multicolumn{2}{|c|}{$1\%$}\\
\cline{3-8}
\multicolumn{2}{|c||}{}&Incremental&proposed&Incremental&proposed&Incremental&proposed\\
\hline
\hline 
 D5 & fraction of ``same sign"& N.A. & 9.99449e-01& N.A. & 9.97822e-01& N.A. & 9.95043e-01\\
 & & &($\pm$1.67e-04)&&($\pm$8.77e-04)&&($\pm$1.16e-03) \\
\cline{2-8}
 & time [sec]& 4.56e-02 & 4.20e-04 & 5.54e-02 & 4.38e-04 & 6.24e-02 & 6.47e-04 \\ 
 & &($\pm$1.23e-02) &($\pm$3.36e-05) &($\pm$1.59e-02) &($\pm$1.90e-05) &($\pm$1.50e-02) &($\pm$3.49e-05)\\ 
\hline
\hline
 D6 & fraction of ``same sign"& N.A. & 1.00000e+00& N.A. & 1.00000e+00& N.A. & 1.00000e+00\\
 & & &($\pm$0.00e+00)&&($\pm$0.00e+00)&&($\pm$0.00e+00) \\
\cline{2-8}
 & time [sec]& 7.37e-02 & 2.70e-03 & 7.02e-02 & 2.55e-03 & 7.54e-02 & 2.90e-03 \\ 
 & &($\pm$1.87e-02) &($\pm$3.38e-04) &($\pm$1.52e-02) &($\pm$1.60e-04) &($\pm$1.47e-02) &($\pm$2.06e-04)\\ 
\hline
\hline
 D7 & fraction of ``same sign"& N.A. & 1.00000e+00& N.A. & 1.00000e+00& N.A. & 1.00000e+00\\
 & & &($\pm$0.00e+00)&&($\pm$0.00e+00)&&($\pm$1.25e-06) \\
\cline{2-8}
 & time [sec]& 6.46e-02 & 4.98e-03 & 6.93e-02 & 4.99e-03 & 7.71e-02 & 5.14e-03 \\ 
 & &($\pm$1.84e-02) &($\pm$3.80e-04) &($\pm$1.61e-02) &($\pm$2.89e-04) &($\pm$1.60e-02) &($\pm$2.40e-04)\\ 
\hline
\hline
 D8 & fraction of ``same sign"& N.A. & 9.99964e-01& N.A. & 9.99952e-01& N.A. & 9.99903e-01\\
 & & &($\pm$2.19e-06)&&($\pm$1.15e-06)&&($\pm$2.55e-06) \\
\cline{2-8}
 & time [sec]& 6.15e+01 & 1.38e-01 & 6.41e+01 & 1.56e-01 & 6.95e+01 & 3.41e-01 \\ 
 & &($\pm$6.23e+00) &($\pm$8.46e-03) &($\pm$1.06e+00) &($\pm$4.81e-03) &($\pm$7.68e+00) &($\pm$1.07e-02)\\ 
\hline
\end{tabular}
\end{scriptsize}
\caption{Results on sensitivity analysis on class labels for various values of $(n_{\rm A}+n_{\rm R})/n_{\rm old}$. The fraction of the test instances whose lower and upper bounds of the decision score have same signs, and the computation time in seconds are listed ($\lambda = 0.1$).}
\label{tab:exp-result10}
\end{center}
\end{table}

\begin{table}[p]
\begin{center}
\begin{scriptsize}
\begin{tabular}{|c|c||c|c||c|c||c|c|}
\hline
\multicolumn{2}{|c||}{}&\multicolumn{6}{|c|}{$(n_{\rm A}+n_{\rm R})/n_{\rm old}$}\\
\cline{3-8}
\multicolumn{2}{|c||}{}&\multicolumn{2}{|c||}{$0.01\%$}&\multicolumn{2}{|c||}{$0.1\%$}&\multicolumn{2}{|c|}{$1\%$}\\
\cline{3-8}
\multicolumn{2}{|c||}{}&Incremental&proposed&Incremental&proposed&Incremental&proposed\\
\hline
\hline 
  D5 & fraction of ``same sign"& N.A. & 1.00000e+00& N.A. & 1.00000e+00& N.A. & 1.00000e+00\\
 & & &($\pm$0.00e+00)&&($\pm$0.00e+00)&&($\pm$0.00e+00) \\
\cline{2-8}
 & time [sec]& 1.51e-02 & 3.86e-04 & 2.01e-02 & 4.18e-04 & 2.25e-02 & 5.54e-04 \\ 
 & &($\pm$8.36e-04) &($\pm$5.01e-06) &($\pm$2.77e-03) &($\pm$3.56e-06) &($\pm$7.19e-04) &($\pm$6.64e-06)\\ 
\hline
\hline
 D6 & fraction of ``same sign"& N.A. & 1.00000e+00& N.A. & 1.00000e+00& N.A. & 1.00000e+00\\
 & & &($\pm$0.00e+00)&&($\pm$0.00e+00)&&($\pm$0.00e+00) \\
\cline{2-8}
 & time [sec]& 2.14e-02 & 2.03e-03 & 2.35e-02 & 2.13e-03 & 2.52e-02 & 2.41e-03 \\ 
 & &($\pm$4.13e-05) &($\pm$3.99e-05) &($\pm$2.31e-03) &($\pm$8.39e-05) &($\pm$4.43e-04) &($\pm$3.49e-05)\\ 
\hline
\hline
 D7 & fraction of ``same sign"& N.A. & 9.99872e-01& N.A. & 9.99869e-01& N.A. & 9.99848e-01\\
 & & &($\pm$4.57e-06)&&($\pm$1.13e-05)&&($\pm$7.02e-06) \\
\cline{2-8}
 & time [sec]& 6.13e-02 & 5.72e-03 & 6.63e-02 & 5.57e-03 & 8.07e-02 & 5.86e-03 \\ 
 & &($\pm$1.08e-02) &($\pm$3.56e-04) &($\pm$1.21e-02) &($\pm$2.49e-04) &($\pm$7.28e-03) &($\pm$2.83e-04)\\ 
\hline
\hline
 D8 & fraction of ``same sign"& N.A. & 9.99925e-01& N.A. & 9.99916e-01& N.A. & 9.99906e-01\\
 & & &($\pm$6.66e-07)&&($\pm$8.98e-07)&&($\pm$3.52e-07) \\
\cline{2-8}
 & time [sec]& 2.60e+01 & 1.40e-01 & 2.90e+01 & 1.53e-01 & 4.02e+01 & 3.44e-01 \\ 
 & &($\pm$1.30e+00) &($\pm$1.01e-02) &($\pm$4.89e+00) &($\pm$4.08e-03) &($\pm$1.18e+00) &($\pm$1.18e-02)\\ 
\hline
\end{tabular}
\end{scriptsize}
\caption{Results on sensitivity analysis on class labels for various values of $(n_{\rm A}+n_{\rm R})/n_{\rm old}$. The fraction of the test instances whose lower and upper bounds of the decision score have same signs, and the computation time in seconds are listed ($\lambda = 1$).}
\label{tab:exp-result11}
\end{center}
\end{table}

\begin{table}[p]
\begin{center}
\begin{scriptsize}
\begin{tabular}{|c|c||c|c||c|c||c|c|}
\hline
\multicolumn{2}{|c||}{}&\multicolumn{6}{|c|}{$n_{\rm old}/n_{\rm train}$}\\
\cline{3-8}
\multicolumn{2}{|c||}{}&\multicolumn{2}{|c||}{$10\%$}&\multicolumn{2}{|c||}{$50\%$}&\multicolumn{2}{|c|}{$99\%$}\\
\cline{3-8}
\multicolumn{2}{|c||}{}&Incremental&proposed&Incremental&proposed&Incremental&proposed\\
\hline
\hline
 D5 & fraction of ``same sign"& N.A. & 9.26417e-01 & N.A. & 9.83689e-01 & N.A. & 9.89409e-01 \\
  & & & ($\pm$4.66e-03)& & ($\pm$3.86e-03)& &($\pm$4.12e-03) \\
\cline{2-8}
 & time [sec]& 8.02e-03 & 4.12e-04& 3.62e-02 & 4.33e-04& 8.14e-02& 4.29e-04\\ 
 & &($\pm$2.32e-05) &($\pm$1.18e-05) &($\pm$1.79e-03) &($\pm$2.26e-05) &($\pm$1.27e-02) &($\pm$1.72e-05)\\ 
\hline
\hline
 D6 & fraction of ``same sign"& N.A. & 1.00000e+00 & N.A. & 1.00000e+00 & N.A. & 1.00000e+00 \\
  & & & ($\pm$0.00e+00)& & ($\pm$0.00e+00)& &($\pm$0.00e+00) \\
\cline{2-8}
 & time [sec]& 1.07e-02 & 2.70e-03& 6.07e-02 & 2.64e-03& 1.04e-01& 2.66e-03\\ 
 & &($\pm$6.79e-05) &($\pm$1.55e-04) &($\pm$1.15e-02) &($\pm$7.42e-05) &($\pm$1.84e-02) &($\pm$1.68e-04)\\ 
\hline
\hline
 D7 & fraction of ``same sign"& N.A. & 9.91657e-01 & N.A. & 9.99136e-01 & N.A. & 9.99477e-01 \\
  & & & ($\pm$5.34e-03)& & ($\pm$2.99e-04)& &($\pm$1.69e-04) \\
\cline{2-8}
 & time [sec]& 1.95e-02 & 6.54e-03& 1.36e-01 & 6.19e-03& 2.02e-01& 6.36e-03\\ 
 & &($\pm$1.13e-03) &($\pm$6.54e-04) &($\pm$3.16e-02) &($\pm$5.76e-04) &($\pm$2.81e-02) &($\pm$5.60e-04)\\ 
\hline
\hline
 D8 & fraction of ``same sign"& N.A. & 9.94789e-01 & N.A. & 9.99324e-01 & N.A. & 9.99582e-01 \\
  & & & ($\pm$5.11e-04)& & ($\pm$3.60e-05)& &($\pm$1.54e-05) \\
\cline{2-8}
 & time [sec]& 2.54e+01 & 1.54e-01& 9.25e+01 & 1.57e-01& 1.39e+02& 1.54e-01\\ 
 & &($\pm$8.29e-01) &($\pm$7.58e-03) &($\pm$9.54e+00) &($\pm$6.73e-03) &($\pm$1.75e+01) &($\pm$5.01e-03)\\ 
\hline
\end{tabular}
\end{scriptsize}
\caption{Results on sensitivity analysis on class labels for various values of $n_{\rm old}/n_{\rm train}$. The fraction of the test instances whose lower and upper bounds of the decision score have same signs, and the computation time in seconds are listed ($\lambda = 0.01$).}
\label{tab:exp-result12}
\end{center}
\end{table}

\begin{table}[p]
\begin{center}
\begin{scriptsize}
\begin{tabular}{|c|c||c|c||c|c||c|c|}
\hline
\multicolumn{2}{|c||}{}&\multicolumn{6}{|c|}{$n_{\rm old}/n_{\rm train}$}\\
\cline{3-8}
\multicolumn{2}{|c||}{}&\multicolumn{2}{|c||}{$10\%$}&\multicolumn{2}{|c||}{$50\%$}&\multicolumn{2}{|c|}{$99\%$}\\
\cline{3-8}
\multicolumn{2}{|c||}{}&Incremental&proposed&Incremental&proposed&Incremental&proposed\\
\hline
\hline
 D5 & fraction of ``same sign"& N.A. & 9.82679e-01 & N.A. & 9.96878e-01 & N.A. & 9.97818e-01 \\
  & & & ($\pm$5.55e-16)& & ($\pm$8.31e-04)& &($\pm$6.69e-04) \\
\cline{2-8}
 & time [sec]& 6.01e-03 & 4.63e-04& 3.02e-02 & 4.42e-04& 5.33e-02& 4.39e-04\\ 
 & &($\pm$5.17e-05) &($\pm$5.10e-05) &($\pm$7.00e-03) &($\pm$1.61e-05) &($\pm$1.61e-02) &($\pm$2.81e-05)\\ 
\hline
\hline
 D6 & fraction of ``same sign"& N.A. & 1.00000e+00 & N.A. & 1.00000e+00 & N.A. & 1.00000e+00 \\
  & & & ($\pm$0.00e+00)& & ($\pm$0.00e+00)& &($\pm$0.00e+00) \\
\cline{2-8}
 & time [sec]& 9.40e-03 & 2.62e-03& 4.33e-02 & 2.60e-03& 7.80e-02& 2.64e-03\\ 
 & &($\pm$7.60e-05) &($\pm$1.55e-04) &($\pm$9.01e-03) &($\pm$1.33e-04) &($\pm$1.10e-02) &($\pm$1.30e-04)\\ 
\hline
\hline
 D7 & fraction of ``same sign"& N.A. & 9.99998e-01 & N.A. & 9.99998e-01 & N.A. & 1.00000e+00 \\
  & & & ($\pm$3.26e-06)& & ($\pm$1.84e-06)& &($\pm$0.00e+00) \\
\cline{2-8}
 & time [sec]& 1.63e-02 & 4.81e-03& 4.49e-02 & 4.84e-03& 7.20e-02& 4.97e-03\\ 
 & &($\pm$5.24e-04) &($\pm$1.92e-04) &($\pm$6.80e-03) &($\pm$2.65e-04) &($\pm$1.51e-02) &($\pm$2.62e-04)\\ 
\hline
\hline
 D8 & fraction of ``same sign"& N.A. & 9.99756e-01 & N.A. & 9.99940e-01 & N.A. & 9.99951e-01 \\
  & & & ($\pm$1.59e-05)& & ($\pm$3.20e-06)& &($\pm$8.67e-07) \\
\cline{2-8}
 & time [sec]& 1.38e+01 & 1.56e-01& 4.20e+01 & 1.55e-01& 6.59e+01& 1.54e-01\\ 
 & &($\pm$2.57e-01) &($\pm$1.11e-02) &($\pm$4.04e-01) &($\pm$5.68e-03) &($\pm$2.39e+00) &($\pm$6.09e-03)\\ 
\hline
\end{tabular}
\end{scriptsize}
\caption{Results on sensitivity analysis on class labels for various values of $n_{\rm old}/n_{\rm train}$. The fraction of the test instances whose lower and upper bounds of the decision score have same signs, and the computation time in seconds are listed ($\lambda = 0.1$).}
\label{tab:exp-result13}
\end{center}
\end{table}

\begin{table}[p]
\begin{center}
\begin{scriptsize}
\begin{tabular}{|c|c||c|c||c|c||c|c|}
\hline
\multicolumn{2}{|c||}{}&\multicolumn{6}{|c|}{$n_{\rm old}/n_{\rm train}$}\\
\cline{3-8}
\multicolumn{2}{|c||}{}&\multicolumn{2}{|c||}{$10\%$}&\multicolumn{2}{|c||}{$50\%$}&\multicolumn{2}{|c|}{$99\%$}\\
\cline{3-8}
\multicolumn{2}{|c||}{}&Incremental&proposed&Incremental&proposed&Incremental&proposed\\
\hline
\hline
 D5 & fraction of ``same sign"& N.A. & 1.00000e+00 & N.A. & 1.00000e+00 & N.A. & 1.00000e+00 \\
  & & & ($\pm$0.00e+00)& & ($\pm$0.00e+00)& &($\pm$0.00e+00) \\
\cline{2-8}
 & time [sec]& 3.09e-03 & 3.83e-04& 1.35e-02 & 4.01e-04& 1.73e-02& 4.13e-04\\ 
 & &($\pm$1.50e-05) &($\pm$6.77e-06) &($\pm$2.12e-03) &($\pm$7.79e-06) &($\pm$2.21e-03) &($\pm$4.42e-06)\\ 
\hline
\hline
 D6 & fraction of ``same sign"& N.A. & 1.00000e+00 & N.A. & 1.00000e+00 & N.A. & 1.00000e+00 \\
  & & & ($\pm$0.00e+00)& & ($\pm$0.00e+00)& &($\pm$0.00e+00) \\
\cline{2-8}
 & time [sec]& 7.36e-03 & 1.86e-03& 1.63e-02 & 2.17e-03& 2.78e-02& 2.22e-03\\ 
 & &($\pm$3.23e-05) &($\pm$1.46e-05) &($\pm$6.71e-04) &($\pm$8.19e-05) &($\pm$9.21e-04) &($\pm$4.33e-05)\\ 
\hline
\hline
 D7 & fraction of ``same sign"& N.A. & 9.99754e-01 & N.A. & 9.99885e-01 & N.A. & 9.99865e-01 \\
  & & & ($\pm$3.28e-05)& & ($\pm$1.50e-05)& &($\pm$6.89e-06) \\
\cline{2-8}
 & time [sec]& 1.66e-02 & 6.38e-03& 5.31e-02 & 5.93e-03& 7.31e-02& 5.86e-03\\ 
 & &($\pm$9.04e-04) &($\pm$3.57e-04) &($\pm$1.00e-02) &($\pm$7.21e-04) &($\pm$1.28e-02) &($\pm$7.41e-04)\\ 
\hline
\hline
 D8 & fraction of ``same sign"& N.A. & 9.99735e-01 & N.A. & 9.99915e-01 & N.A. & 9.99916e-01 \\
  & & & ($\pm$2.19e-05)& & ($\pm$1.49e-06)& &($\pm$8.67e-07) \\
\cline{2-8}
 & time [sec]& 6.87e+00 & 1.48e-01& 2.36e+01 & 1.56e-01& 2.83e+01& 1.52e-01\\ 
 & &($\pm$8.04e-01) &($\pm$1.05e-02) &($\pm$3.73e+00) &($\pm$6.65e-03) &($\pm$2.54e+00) &($\pm$3.29e-03)\\ 
\hline
\end{tabular}
\end{scriptsize}
\caption{Results on sensitivity analysis on class labels for various values of $n_{\rm old}/n_{\rm train}$. The fraction of the test instances whose lower and upper bounds of the decision score have same signs, and the computation time in seconds are listed ($\lambda = 1$).}
\label{tab:exp-result14}
\end{center}
\end{table}

\begin{table*}[p]
\begin{center}
\begin{scriptsize}
\begin{tabular}{|c|c||c|c|c||c|c||c||c|c|}
\hline
\multicolumn{2}{|c||}{}&\multicolumn{5}{|c||}{SVM}&\multicolumn{3}{|c|}{Logistic regression}\\
\cline{3-10}
\multicolumn{2}{|c||}{}&\multicolumn{3}{|c||}{existing}&\multicolumn{2}{|c||}{proposed}&existing&\multicolumn{2}{|c|}{proposed}\\
\cline{3-10}
\multicolumn{2}{|c||}{}&incremental&\cite{Vapnik96}&\cite{Joachims2000}&op1&op2&incremental&op1&op2\\
\hline
\hline
D1&linear&13.76&10.46&10.37&7.74&5.17&13.95&8.32&2.78\\
\cline{2-10}
&nonlinear&33.19&24.53&15.63&17.33&8.48&29.31&17.23&6.55\\
\hline
\hline
D2&linear&52.87&51.04&44.28&29.01&15.93&58.22&28.63&13.25\\
\cline{2-10}
&nonlinear&337.87&312.44&246.10&201.58&124.79&268.71&165.42&120.57\\
\hline
\hline
D3&linear&1167.65&458.12&229.57&203.60&123.69&4075.96&726.68&346.30\\
\cline{2-10}
&nonlinear&96317.45&77562.37&46427.27&58480.90&46301.53&91503.32&34972.51&28856.92\\
\hline
\hline
D4&linear&18824.74&14303.27&12177.41&8506.30&2088.63&25197.11&10563.92&1219.36\\
\cline{2-10}
&nonlinear&$>$ 3 days&$>$ 3 days&183208.76&202972.55&106169.25&$>$ 3 days&125300.26&47474.64\\
\hline
\end{tabular}
\end{scriptsize}
\caption{Computation time [sec] of model selection based on LOOCV (without tricks).}
\label{tab:exp-result15}
\end{center}
\end{table*}

\begin{table*}[p]
\begin{center}
\begin{scriptsize}
\begin{tabular}{|c|c||c|c|c||c|c||c||c|c|}
\hline
\multicolumn{2}{|c||}{}&\multicolumn{5}{|c||}{SVM}&\multicolumn{3}{|c|}{Logistic regression}\\
\cline{3-10}
\multicolumn{2}{|c||}{}&\multicolumn{3}{|c||}{existing}&\multicolumn{2}{|c||}{proposed}&existing&\multicolumn{2}{|c|}{proposed}\\
\cline{3-10}
\multicolumn{2}{|c||}{}&incremental&\cite{Vapnik96}&\cite{Joachims2000}&op1&op2&incremental&op1&op2\\
\hline
\hline
D1&linear&8.88&8.69&8.53&5.79&4.45&5.41&3.96&1.66\\
\cline{2-10}
&nonlinear&14.59&13.13&10.26&9.00&4.50&12.32&7.42&2.54\\
\hline
\hline
D2&linear&17.88&17.88&16.65&1.44&0.83&20.49&1.20&0.55\\
\cline{2-10}
&nonlinear&164.39&151.57&138.15&106.15&47.66&125.23&76.95&44.44\\
\hline
\hline
D3&linear&693.34&345.10&226.65&197.68&124.81&2012.49&563.09&322.47\\
\cline{2-10}
&nonlinear&9018.88&5805.36&4772.59&1898.11&1352.38&8495.91&1184.83&745.02\\
\hline
\hline
D4&linear&6132.45&5536.21&4121.21&353.21&93.67&12027.28&663.19&187.13\\
\cline{2-10}
&nonlinear&168806.92&139810.43&122264.81&46166.76&23032.34&143660.82&35676.66&14920.24\\
\hline
\end{tabular}
\end{scriptsize}
\caption{Computation time [sec] of model selection based on LOOCV (with tricks).}
\label{tab:exp-result16}
\end{center}
\end{table*}

\end{document}